\pdfoutput=1
\documentclass[journal]{IEEEtai}

\newif\ifanonymous
\anonymoustrue

\newif\ifarxiv
\arxivtrue
\ifarxiv\anonymousfalse\fi

\newif\ifshownotes
\shownotesfalse

\usepackage{amsmath,amssymb,amsfonts}
\usepackage{graphicx}
\usepackage{array}
\usepackage{booktabs}
\usepackage{multirow}
\usepackage{xcolor}
\usepackage{cite}                 
\usepackage{algorithm}
\usepackage{algpseudocode}
\usepackage{bm}
\usepackage{url}
\usepackage[caption=false,font=footnotesize]{subfig}
\graphicspath{{figures/}}

\usepackage[colorlinks,urlcolor=blue,linkcolor=blue,citecolor=blue]{hyperref}

\ifshownotes
  
  \newcommand{\todo}[1]{{\color{orange}[TODO: #1]}}
\else
  
  \newcommand{\todo}[1]{}
\fi

\newcommand{\method}{KiRA}
\newcommand{\kron}{Kronecker}

\newcommand{\R}{\mathbb{R}}
\newcommand{\dW}{\Delta W}
\newcommand{\kr}{\otimes}
\DeclareMathOperator{\rank}{rank}
\DeclareMathOperator{\erank}{erank}
\DeclareMathOperator{\srank}{srank}
\DeclareMathOperator{\diag}{diag}
\DeclareMathOperator{\vect}{vec}
\newcommand{\norm}[1]{\left\lVert #1 \right\rVert}

\newtheorem{proposition}{Proposition}

\newtheorem{lemma}{Lemma}

\begin{document}

 \title{KiRA: Kronecker Input-Gated Rank Adaptation for Parameter-Efficient Fine-Tuning of Pretrained Models}

\author{Akil Ahmad Taki ,  Shaikh Anowarul Fattah\\[2pt]
{\normalsize Department of Electrical and Electronic Engineering}\\
{\normalsize Bangladesh University of Engineering and Technology (BUET),
Dhaka 1205, Bangladesh}\\[1pt]
{\normalsize\ttfamily 1806139@eee.buet.ac.bd, fattah@eee.buet.ac.bd}}
\markboth{}{}

\maketitle

\maketitle

\begin{abstract}
Full fine-tuning of large pretrained models is expensive, and
parameter-efficient fine-tuning (PEFT) is the standard alternative. Most
PEFT methods use low-rank updates, whose rank lies far below that of full
fine-tuning and is the same for every input. Input-adaptive variants vary
the capacity per input but remain within the low-rank bottleneck, and
whether a task benefits from input conditioning is not known before
training. This paper introduces Kronecker input-gated rank adaptation
(KiRA). Instead of the conventional low-rank product, the update $\Delta W$
is the Kronecker product of a frozen random $p\times p$ matrix $A$ and a
trainable factor $M(x)$, which overcomes the rank ceiling: because rank is
multiplicative under this product, $A$ multiplies the rank of $M(x)$ by
$p$ without adding trainable parameters. The factor $M(x)$ is further
decomposed as $B_1\,\mathrm{diag}(g(x))\,B_2$, where the input gate
$g(x)$, computed once per sample by a network shared across all adapted
layers, adjusts the rank of the update to each input in steps of $p$.
Where input conditioning is unnecessary, the gate is folded into $B_1$ at
deployment. Because the gate changes the rank only in discrete steps, the
benefit of input conditioning can be predicted before training from a few
hundred labeled inputs. Across six medical imaging and language
benchmarks, KiRA attains the highest mean score against five published
baselines with 8.7--9.7 times fewer trainable parameters than low-rank
adaptation, and the pre-training criterion agrees with the measured
benefit of input conditioning on all six tasks.
\end{abstract}

\begin{IEEEImpStatement}
Large pretrained models are increasingly adapted to many specialized tasks
across healthcare, language technology, and other domains, and each
adaptation adds storage, memory, and energy costs that limit where these
models can be deployed. This work shows that the trainable part of an
adaptation can be made substantially smaller without loss of accuracy, by
pairing a small trainable module with a fixed matrix
that multiplies the module's capacity without adding parameters to train. Smaller adaptations make it practical for hospitals, small
organizations, and devices with limited memory to maintain many
task-specific models on shared hardware. The work also provides a
criterion, measured on a small set of labeled examples before training,
that indicates whether a more complex input-dependent design is worth its
cost, which reduces wasted computation in model development. The results
were obtained on public benchmarks.
\end{IEEEImpStatement}




\section{Introduction}\label{sec:intro}

\IEEEPARstart{A}{dapting} a large pretrained model to a new task by
updating all of its weights is often impractical, particularly when a
single foundation model serves many clinical tasks~\cite{jiao2026medfm}
or must operate on resource-constrained devices~\cite{dritsas2026edge}.
Parameter-efficient methods therefore freeze the model and learn a small
additive update to selected weight matrices. Low-rank
adaptation~\cite{hu2022lora} is the standard approach: the update is the
product of two thin matrices, so that its rank is bounded by a chosen
constant $r$. Although highly effective, this approach rests on two
assumptions: that an update of rank $r$ is sufficient, and that the same
rank is appropriate for every input.

Prior work addresses each assumption separately. One line of work
increases or redistributes the rank available under a fixed parameter
budget, by allocating rank across weight matrices~\cite{zhang2023adalora},
reparameterizing the update~\cite{liu2024dora}, freezing random
projections~\cite{kopiczko2024vera}, or replacing the low-rank product with
a Kronecker product~\cite{edalati2022krona}. Another line makes the update
input-dependent, with a small network selecting the directions of the
update that are used for each input~\cite{wu2024mole,li2026flexilora}.
The second line has been developed mainly on a low-rank substrate, on
which activating or deactivating one direction changes the rank by one.

Two questions remain open. Input-adaptive adapters require additional
parameters and computation, and no criterion indicates in advance whether
a given task will benefit from them. Moreover, comparisons with static
adapters do not separate the two contributions of a gate: access to the
input, and the additional parameters of the gating network.

This paper addresses both questions. We propose Kronecker input-gated rank
adaptation (\method{}), whose update is the Kronecker product of a frozen
random factor and a gated trainable factor. Because the rank of a
Kronecker product is the product of the ranks of its factors, the frozen
factor multiplies the rank of the trainable factor by its size $p$ without
adding trainable parameters, and each gate unit controls $p$ units of rank
rather than one. The same identity determines the resolution of the gate:
the rank of the update varies in discrete steps whose size is set by $p$,
so that input conditioning can be effective only on tasks whose per-input
demand varies by more than one step. This condition can be evaluated on
the pretrained model before training.

To test these predictions, a factorial design crosses the substrate
(low-rank or Kronecker) with the adaptivity (static or input-gated) on six
tasks in two modalities. Each gated configuration is paired with a
constant-gate control that retains the parameters of the gate but not its
access to the input, so that the gain from gating is decomposed rather
than attributed to adaptivity by assumption. The main contributions of
this paper are as follows:
\begin{enumerate}
\item \textbf{Gated Kronecker adapter.} We propose \method{}, an adapter
  whose update is the Kronecker product of a frozen random factor and a
  trainable low-rank factor with an input-conditioned diagonal gate. The
  frozen factor multiplies the rank of the trainable factor by $p$ without
  adding trainable parameters (Section~\ref{sec:method:form}).
\item \textbf{Exact rank identities and amplification trade-off.} We
  derive exact rank identities for the update, $\rank(\dW(x))=p\,k(x)$ for
  a hard gate and multiplicative effective and stable ranks for a soft
  gate, and show that amplification reduces the dimension of the reachable
  subspace (Sections~\ref{sec:method:rank}--\ref{sec:method:tradeoff}).
\item \textbf{Pre-training criterion for input conditioning.} We show that
  the gate changes the rank only in steps of the effective rank of the
  frozen factor, and derive from this an adaptivity ratio, computed on a
  few hundred labeled training inputs before training, together with a
  criterion that selects whether the gate should depend on the input
  (Sections~\ref{sec:method:lattice}--\ref{sec:method:criterion}).
\item \textbf{Controlled empirical evaluation.} On six medical imaging and
  language tasks, \method{} attains the highest mean score on every task
  against five published baselines, with $8.7$--$9.7$ times fewer
  trainable parameters than low-rank adaptation. Parameter-matched
  constant-gate controls show that the accuracy advantage derives from the
  Kronecker substrate, and that input conditioning contributes only on the
  three tasks that the criterion selects (Section~\ref{sec:results}).
\end{enumerate}


\section{Background and Related Work}\label{sec:background}

Adapters freeze a pretrained model and train a small module
alongside it~\cite{houlsby2019adapters}. Low-rank
adaptation~\cite{hu2022lora} expresses the update as $\dW=BA$ with inner
dimension $r$, motivated by evidence that fine-tuning objectives have a
low intrinsic dimension~\cite{li2018intrinsic,aghajanyan2021intrinsic}.
The rank $r$ is fixed before training and is the same for every input.

One line of work retains the low-rank form and distributes the parameter
budget more effectively. AdaLoRA parameterizes the update in the form of a
singular value decomposition and prunes the singular values of less
important weight matrices, reallocating the budget among
them~\cite{zhang2023adalora}; DyLoRA trains a range of ranks so that one
can be selected after training~\cite{valipour2023dylora}; DoRA decomposes
the pretrained weight into magnitude and direction and applies the
low-rank update to the direction only~\cite{liu2024dora}; VeRA freezes a
pair of random projections and trains only scaling vectors, which reduces
the parameter count at a fixed rank~\cite{kopiczko2024vera}; and ARD-LoRA
learns per-head ranks through differentiable scaling factors under a
sparsity and smoothness objective~\cite{shinwari2025ardlora}. These methods
differ in how the budget is distributed, but in each of them the
allocation is fixed at the end of training, so that every input receives
an update of the same rank at inference.

A second line of work changes the factorization rather than the budget.
KronA replaces the low-rank product with a Kronecker
product~\cite{edalati2022krona}, whose rank is the product of the ranks of
its factors, and subsequent work combines Kronecker and low-rank
factors~\cite{shen2025kronlora} or studies the design of their
components~\cite{bai2026cdka}. In these methods, both factors are
trained. \method{} instead keeps one factor frozen and random, so that the
amplification of rank requires no trainable parameters. MoKA combines
several Kronecker adapters with different factor shapes through learned
mixture weights~\cite{sadeghi2025moka}. Its weights act on entire
factorizations, whereas the gate of \method{} modulates the inner
directions of a single factorization, which yields the exact rank identity
of Section~\ref{sec:method:rank}.

Input-dependent computation is established in conditional computation and
mixture-of-experts routing~\cite{shazeer2017moe,fedus2022switch}, with
recent applications to domain adaptation~\cite{du2025moeosda} and to
efficient transformer inference~\cite{jaradat2025hdp}. In parameter-efficient
fine-tuning, it has been applied by combining low-rank experts per
input~\cite{wu2024mole} and by selecting the rank of a low-rank update for
each sample through a learned router~\cite{li2026flexilora}. These methods
operate on a low-rank substrate, and their evaluations compare the gated
adapter with a static one, which does not separate the effect of access to
the input from that of the additional parameters of the routing network.

Two elements are therefore not addressed in prior comparisons. First, a
gated adapter differs from its static counterpart in two respects: it
depends on the input, and it introduces the parameters of the gating
network. The gain from gating cannot be attributed to adaptivity without a
control that retains the second difference and removes the first; such a
control is added to each substrate in Section~\ref{sec:setup}. Second, to
our knowledge, no existing criterion predicts before training which tasks
benefit from input conditioning. The criterion derived in
Section~\ref{sec:method:criterion} depends only on the task and the
adapter configuration and is measured on the pretrained model.


\section{Methodology}\label{sec:method}

\method{} is designed to satisfy two requirements. First, each trainable
parameter must contribute more rank to the update than it does in a
low-rank adapter. Second, rank must be allocated according to the input
only on tasks for which this allocation is predicted, before training, to
improve performance. A frozen random factor satisfies the first
requirement by supplying rank without trainable parameters, and an
input-conditioned gate whose benefit can be assessed on the pretrained
model satisfies the second. Section~\ref{sec:method:form} defines the
update, Sections~\ref{sec:method:rank}--\ref{sec:method:lattice} derive its
properties, and Section~\ref{sec:method:criterion} presents the
pre-training criterion.

\subsection{From low-rank to gated Kronecker updates}\label{sec:method:form}

\begin{figure}[!h]
  \centering
  \includegraphics[width=0.82\columnwidth]{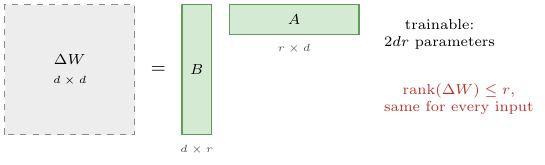}
  \caption{Low-rank adaptation. The update $\dW=BA$ is the product of two
  thin trainable matrices, so its rank is at most $r$ for every input.}
  \label{fig:lora}
\end{figure}

A transformer layer applies dense weight matrices to its input. Let
$W_0\in\R^{d\times d}$ be one such pretrained projection inside an attention
block, and let $x\in\R^{d}$ be a token representation. During fine-tuning,
an adapter keeps $W_0$ frozen and learns an update $\dW$, so that the
layer computes
\begin{equation}\label{eq:forward}
  y \;=\; W_0x + \dW x .
\end{equation}
Low-rank adaptation~\cite{hu2022lora} represents the update as $\dW=BA$,
with $B\in\R^{d\times r}$, $A\in\R^{r\times d}$, and $r\ll d$
(Fig.~\ref{fig:lora}). The rank of $\dW$ is therefore at most $r$ for
every input, and a higher rank requires a larger $r$, which increases the
number of trainable parameters proportionally.

\begin{figure}[!h]
  \centering
  \includegraphics[width=0.94\columnwidth]{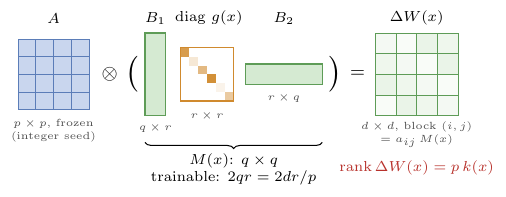}
  \caption{The \method{} update \eqref{eq:main}. The frozen factor $A$ is
  combined by a Kronecker product with the trainable factor $M(x)$ of
  \eqref{eq:mfactor}. Block $(i,j)$ of $\dW(x)$ equals $a_{ij}M(x)$, so
  each direction that the gate activates in $M(x)$ is repeated in all $p$
  block rows.}
  \label{fig:kron}
\end{figure}

\method{} replaces the product $BA$ by a Kronecker product of two factors
(Fig.~\ref{fig:kron}):
\begin{equation}\label{eq:main}
  \dW(x) \;=\; A \kr M(x).
\end{equation}
Here $A\in\R^{p\times p}$, unlike the trainable factor $A$ of low-rank
adaptation, is a random matrix with independent standard Gaussian entries
that is drawn once and kept frozen; only the integer seed that generates it
is stored. The second factor
$M(x)\in\R^{q\times q}$, with $q=d/p$, is trainable and depends on the
input. Since the Kronecker product of a $p\times p$ and a $q\times q$
matrix has size $pq\times pq=d\times d$, $p$ must be a divisor of $d$.

The trainable factor is a gated low-rank product,
\begin{equation}\label{eq:mfactor}
  M(x) \;=\; B_1 \diag\bigl(g(x)\bigr) B_2 ,
\end{equation}
where $B_1\in\R^{q\times r}$ and $B_2\in\R^{r\times q}$ are trainable, and
$g(x)=(g_1(x),\dots,g_r(x))\in[0,1]^{r}$ is a vector of $r$ gate values,
one for each column of $B_1$ and the corresponding row of $B_2$. Each
adapted layer is specified by the pair $(p,r)$, and the forward pass
\eqref{eq:forward} becomes $y=W_0x+\dW(x)\,x$.

The frozen factor increases the rank of the update without adding
trainable parameters. Since $\rank(X\kr Y)=\rank(X)\rank(Y)$, and a random
square matrix has full rank with probability one, the rank of $\dW(x)$ is
$p$ times the rank of $M(x)$, which is at most $r$. The trainable
parameters of the update are contained in $B_1$ and $B_2$, which together
have $2qr=2dr/p$ entries per adapted layer. Increasing $p$ therefore
increases the attainable rank and decreases the number of trainable
parameters at the same time; Section~\ref{sec:method:tradeoff} describes
the corresponding cost.

Keeping $A$ frozen and random is supported by both the analysis and prior
work. The rank identities of Section~\ref{sec:method:rank} require only
that $A$ has full rank and a spread spectrum. A Gaussian matrix has full
rank with probability one and an effective rank of approximately $0.80p$
(Section~\ref{sec:method:rank}), so that the amplification is obtained
without data or training. Training $A$ would instead add $p^{2}$ parameters
per adapted layer, $49{,}152$ in total at $p=32$, which exceeds the
complete vision substrate by a factor of four. Frozen random matrices are
also established in fine-tuning: optimization within a fixed random
subspace suffices for many objectives~\cite{li2018intrinsic,aghajanyan2021intrinsic},
random projections or random bases with trained coefficients match
low-rank adaptation with far fewer trainable
parameters~\cite{kopiczko2024vera,koohpayegani2024nola}, and the
projection matrix of a low-rank adapter can remain at its random
initialization with little loss in
accuracy~\cite{zhang2023lorafa,zhu2024asymmetry}.

\begin{figure}[!h]
  \centering
  \includegraphics[width=0.94\columnwidth]{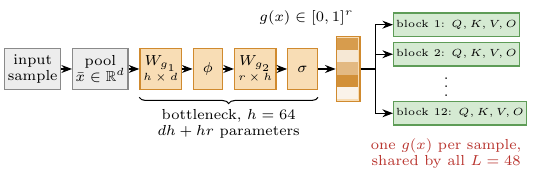}
  \caption{The gate \eqref{eq:gate}. One bottleneck network maps the
  mean-pooled representation $\bar{x}$ of the input sample to
  $g(x)\in[0,1]^{r}$, computed once per sample and shared by all $L$
  adapted projections.}
  \label{fig:gate}
\end{figure}

The gate makes the rank depend on the input. A small network, shared by
all $L$ adapted projections, maps the mean-pooled token representation
$\bar{x}\in\R^{d}$ of the input sample to the gate values
(Fig.~\ref{fig:gate}):
\begin{equation}\label{eq:gate}
  g(x) \;=\; \sigma\!\left(W_{g_2}\,\phi\!\left(W_{g_1}\bar{x}\right)\right),
\end{equation}
where $W_{g_1}\in\R^{h\times d}$ and $W_{g_2}\in\R^{r\times h}$ are
trainable weights, $\phi$ is the GELU nonlinearity, $\sigma$ is the
logistic function, and $h=64$. The gate is computed once per input sample,
an image or a sentence pair, and the same vector $g(x)$ is used in every
adapted layer; it therefore adds $h(d+r)$ trainable parameters regardless
of the number of layers. Because $A$ repeats each direction of $M(x)$
across $p$ blocks of $\dW(x)$, switching one gate value on or off changes
the rank of the update by $p$ rather than by one. This property
distinguishes \eqref{eq:main} from a gated low-rank adapter.

\subsection{Rank and spectral amplification}\label{sec:method:rank}

The rank of the update in \eqref{eq:main} is not only bounded but
determined exactly: algebraically for a hard gate, and spectrally for the
soft gate used in practice.

\begin{proposition}\label{prop:rank}
Let $A\in\R^{p\times p}$ have full rank, and let the gate be hard, i.e.,
each entry $g_i(x)\in\{0,1\}$, with $k(x)=\sum_{i=1}^{r}g_i(x)$ the number
of active gate units. If the $k(x)$ columns of $B_1$ and the $k(x)$ rows of
$B_2$ selected by $g(x)$ are linearly independent, then
\begin{equation}\label{eq:rank}
  \rank\bigl(\dW(x)\bigr) \;=\; p\,k(x),
  \qquad 0\le k(x)\le r .
\end{equation}
\end{proposition}

The proof follows from $\rank(X\kr Y)=\rank(X)\rank(Y)$, since the
independence condition implies $\rank(M(x))=k(x)$. The condition fails
only on a set of parameters of Lebesgue measure zero and therefore holds
almost surely under random initialization. Full proofs of the statements
in this section are given in the supplementary material.

Consequently, the realized rank varies with the input in increments of
$p$, whereas a gate applied to a low-rank adapter yields
$\rank(\dW(x))=k(x)\le r$, an increment of one. The maximum rank $pr$ is
attained although $A$ contains no trainable parameters.

\subsubsection{Soft gates}
In practice, each entry $g_i(x)$ lies in the open interval $(0,1)$. Since
no entry is exactly zero, \eqref{eq:rank} gives $\rank(\dW(x))=pr$ for
every input, and the algebraic rank no longer distinguishes between
inputs. The distinction, together with the amplification, persists in the
spectrum. For a nonzero matrix $X$ with singular values
$\sigma_1\ge\dots\ge\sigma_n$ and $\pi_i=\sigma_i/\sum_j\sigma_j$, the
effective rank~\cite{roy2007erank} and the stable
rank~\cite{rudelson2007sampling} are defined as
\begin{equation}\label{eq:eranks}
  \erank(X)=\exp\Bigl(-\sum_i \pi_i\log\pi_i\Bigr),
  \qquad
  \srank(X)=\frac{\norm{X}_F^{2}}{\norm{X}_2^{2}} .
\end{equation}
Both are continuous measures of rank that count the directions carrying
appreciable energy rather than all nonzero directions.

\begin{proposition}\label{prop:spectral}
For any nonzero $A$ and $M(x)$,
\begin{equation}\label{eq:mult}
  \begin{aligned}
  \erank\bigl(A\kr M(x)\bigr)&=\erank(A)\,\erank\bigl(M(x)\bigr),\\
  \srank\bigl(A\kr M(x)\bigr)&=\srank(A)\,\srank\bigl(M(x)\bigr).
  \end{aligned}
\end{equation}
\end{proposition}

Both identities are exact. The singular values of a Kronecker product are
the pairwise products $\sigma_i\tau_j$; hence the spectral entropy is
additive, and the Frobenius and spectral norms are multiplicative. A soft
gate therefore provides the same amplification as a hard gate, with
$\erank(M(x))$ in place of $k(x)$.

\subsubsection{Amplification factor of a Gaussian frozen factor}
The amplification factor in \eqref{eq:mult} is $\erank(A)$ rather than
$p$. The two coincide only when all singular values of $A$ are equal,
which does not hold for a Gaussian $A$.
As $p\to\infty$, the singular values of $A/\sqrt{p}$ follow the
quarter-circle law on $[0,2]$ with density
$\sqrt{4-s^{2}}/\pi$~\cite{marchenko1967}. Substituting this density into
\eqref{eq:eranks} gives
\begin{equation}\label{eq:alphalimit}
  \frac{\erank(A)}{p}\;\longrightarrow\;
  \exp\!\Bigl(-\!\int_{0}^{2}\!\frac{\sqrt{4-s^{2}}}{\pi}\,
  \frac{s}{\bar{s}}\log\frac{s}{\bar{s}}\,ds\Bigr)\approx0.805,
\end{equation}
where $\bar{s}=8/(3\pi)$ is the mean of the limiting distribution, and
likewise $\srank(A)/p\to1/4$. The limit is approached rapidly: over $2000$
draws, $\erank(A)/p$ ranges from $0.787$ at $p=8$ to $0.803$ at $p=128$
and equals $0.799\pm0.010$ at $p=32$ (Table~S5 of the supplementary
material). The analysis therefore uses the amplification factor
\begin{equation}\label{eq:alpha}
  \alpha \;\triangleq\; \erank(A) \;\approx\; 0.80\,p
\end{equation}
instead of assuming $\alpha=p$. Drawing $A$ uniformly from the orthogonal
group would equalize the singular values and give $\alpha=p$; this variant
is not evaluated here.

\subsection{Trade-off between amplification and reachable dimension}
\label{sec:method:tradeoff}

The amplification provided by $A$ has a cost. For a fixed $A$, the update
is confined to the linear subspace
\begin{equation}\label{eq:reach}
  \mathcal{A} \;=\; \bigl\{\,A\kr M \;:\; M\in\R^{q\times q}\bigr\}
  \;\subset\; \R^{d\times d},
\end{equation}
whose dimension is $D=q^{2}$, independent of $r$ and of the gate.
Directions of $\R^{d\times d}$ outside $\mathcal{A}$ cannot be reached by
any choice of $B_1$, $B_2$, and $g$. Since $q=d/p$, every divisor $p$ of
$d$ satisfies
\begin{equation}\label{eq:tradeoff}
  p^{2}D \;=\; d^{2}.
\end{equation}
The product $p^{2}D$ is thus fixed by the layer width, and amplification
and reachable dimension cannot be increased simultaneously: doubling $p$,
and hence the rank contributed by each gate unit, reduces the reachable
dimension by a factor of four. The cost of a large $p$ is therefore
geometric rather than parametric, since the number of trainable
parameters, $2dr/p$, decreases as $p$ grows. Consequently, $p$ is a
hyperparameter that trades amplification against the dimension of the
reachable subspace, and it cannot be increased without limit.

\subsection{Gate lattice and resolution limit}\label{sec:method:lattice}

Because the gate activates or deactivates inner directions, the rank that
it delivers is quantized. By Propositions~\ref{prop:rank}
and~\ref{prop:spectral}, a hard gate with $k(x)$ active units yields an
algebraic rank of $p\,k(x)$ and an effective rank of at most
$\alpha k(x)$. The realized rank therefore lies, approximately, on the
lattice $\{0,\alpha,2\alpha,\dots,\alpha r\}$ of spacing $\alpha$. The
factor that makes each gate unit contribute $\alpha$ units of rank thus
also makes $\alpha$ the smallest rank difference that the gate can express
between two inputs.

Whether this resolution is sufficient depends on the task. Let $R^{*}(x)$
denote the rank demanded by the objective for input $x$ at a given layer,
and let $\bar{r}^{*}$ and $s^{*}$ denote its mean and standard deviation
over the input distribution. Both quantities are measured on the
pretrained model before any adapter is trained, as described in
Section~\ref{sec:setup}.

\begin{proposition}[Gate resolution]\label{prop:lattice}
Let $\hat{k}(x)$ and $m$ be the elements of $\{0,\dots,r\}$ nearest to
$R^{*}(x)/\alpha$ and $\bar{r}^{*}/\alpha$, respectively, let
$\gamma=s^{*}/\alpha$, and let $\delta=|\bar{r}^{*}-m\alpha|<\alpha/2$.
Then
\begin{equation}\label{eq:cheb}
  \Pr\bigl[\hat{k}(x)\neq m\bigr]
  \;\le\; \frac{\gamma^{2}}{\left(\tfrac12-\delta/\alpha\right)^{2}} .
\end{equation}
\end{proposition}

The proof combines a rounding argument with Chebyshev's inequality and
requires no assumption on the training dynamics. For $\delta=0$, the bound
equals $4\gamma^{2}$: at $\gamma=0.1$, at most $4\%$ of inputs are
assigned a number of gate units different from $m$, and on the remaining
inputs the best input-conditioned gate coincides with the best constant
gate. The bound is vacuous for $\gamma\ge1/2$; the proposition therefore
identifies when gating cannot help, not when it must.

This result motivates the central quantity of the analysis. The
\emph{adaptivity ratio} of a layer and its task-level aggregate are
defined as
\begin{equation}\label{eq:gamma}
  \gamma \;=\; \frac{s^{*}}{\alpha},
  \qquad
  \Gamma \;=\; \frac{1}{L}\bigl|\{\ell : \gamma(\ell)\ge 1\}\bigr| .
\end{equation}
A layer with $\gamma\ge1$ has a demand spread exceeding one lattice step,
so that the gate can distinguish between inputs; for small $\gamma$,
Proposition~\ref{prop:lattice} shows that it cannot. $\Gamma$ can be
computed before training, since $s^{*}$ is obtained from the pretrained
model and $\alpha$ from the chosen $p$.

Two qualifications apply. The condition is necessary but not sufficient,
since the optimizer must still find a gate that exploits the spread; and a
soft gate does not round strictly, so that the transition near $\gamma=1$
is gradual. The resulting prediction is nevertheless testable before
training: on a task with small $\Gamma$, an input-conditioned gate should
perform no better than a constant gate, and the adapter can be deployed as
a static Kronecker adapter.

\subsection{Pre-training criterion for input conditioning}
\label{sec:method:criterion}

Section~\ref{sec:method:lattice} reduces the question of whether input
conditioning can be beneficial to a comparison that is available before
training. Given a candidate value $p_0$, typically the value that would be
used for a static adapter, the gate lattice has spacing
$\alpha_0=\erank(A_{p_0})$, and $\Gamma$ evaluated at $\alpha_0$ gives the
fraction of adapted layers whose demand spread is wide enough for an
input-conditioned gate to distinguish between inputs.
Algorithm~\ref{alg:rule} converts this quantity into a selection between a
gated and a static adapter. Since the transition near $\gamma=1$ is
gradual, the threshold $\Gamma_0=0.5$ is a majority criterion rather than
a sharp boundary; on the six tasks studied here, any value in $(0,0.75]$
yields the same selections (Table~\ref{tab:gamma}).

\begin{algorithm}[!t]
\caption{Selection of input conditioning before training}
\label{alg:rule}
\begin{algorithmic}[1]
\Require pretrained model, a few hundred labeled training inputs,
         candidate $p_0\mid d$, threshold $\Gamma_0=0.5$
\State measure $s^{*}(\ell)$ for each adapted layer (no adapter, no training)
\State $\alpha_0 \gets \erank(A_{p_0})$;\quad
       $\Gamma \gets \frac{1}{L}\,|\{\ell : s^{*}(\ell) \ge \alpha_0\}|$
\If{$\Gamma \ge \Gamma_0$} \Comment{demand spread exceeds the lattice step}
  \State \textbf{gated}: train with the input-conditioned gate $g(x)$
\Else
  \State \textbf{static}: train with a constant gate $c$, then fold
         $B_1 \gets B_1\diag(c)$ and discard the gate network
\EndIf
\end{algorithmic}
\end{algorithm}

Two properties of the criterion should be noted. First, the static branch
replaces the gate with a learned constant rather than removing it. Since
$B_1\diag(c)B_2$ is reachable by $B_1B_2$ alone, a constant gate adds no
expressive power, and once it is folded into $B_1$ it incurs no inference
cost; the selection therefore concerns only whether the gate should depend
on the input. Second, $\Gamma$ depends on $p_0$, because decreasing $p_0$
refines the lattice at the cost of additional parameters. The criterion
does not determine this budget; it determines whether input conditioning
can be effective at the chosen budget.

\subsection{Implementation and cost}\label{sec:method:impl}

The update in \eqref{eq:main} is never formed explicitly. Writing
$x\in\R^{d}$ as a matrix $X\in\R^{q\times p}$ with $\vect(X)=x$, the
identity $(A\kr M)\vect(X)=\vect(MXA^{\top})$ reduces the product to two
small matrix multiplications, and the factored form of $M$ keeps both of
them narrow:
\begin{equation}\label{eq:vectrick}
  \dW(x)\,x \;=\; \vect\!\Bigl(
    B_1\bigl(\diag(g(x))\,(B_2X)\bigr)A^{\top}\Bigr).
\end{equation}
The largest intermediate has size $q\times p$, the size of $x$ itself, so
the $d\times d$ update is never held in memory. The cost is
$2rd+rp+dp$ multiply--accumulate operations per token, compared with $d^{2}$
for a dense update.

Two costs remain. First, the $dp$ term arising from $A^{\top}$ has no
counterpart in a low-rank adapter, whose cost is $2rd$ per token, so the
Kronecker forward pass is more expensive per token. Second, the gate adds
$h(d+r)$ parameters and one forward pass per sample rather than per token,
since $g(x)$ is shared across the $L$ adapted projections. In the static
branch of Algorithm~\ref{alg:rule}, the gate is trained and subsequently
folded into $B_1$, so that it incurs neither parameters nor computation at
inference.

To prevent the gate from collapsing to all units open, which reproduces
the static adapter, or to all units closed, which removes the update, its
mean activity is regularized toward a target $\rho_0$ (values in
Section~\ref{sec:setup}):
\begin{equation}\label{eq:gatereg}
  \mathcal{L} \;=\; \mathcal{L}_{\mathrm{task}}
  \;+\; \lambda\,\mathrm{KL}\bigl(\bar{g}\,\|\,\rho_0\bigr),
  \qquad
  \bar{g}=\tfrac{1}{r}\textstyle\sum_{i} \mathbb{E}_x[g_i(x)].
\end{equation}
During an initial warm-up phase, the gate is held at its neutral value
$g\equiv\rho_0$, so that the magnitude of the update stabilizes before the
gate begins to distribute it across inputs. The penalty is symmetric: it
discourages both extremes without favoring sparse gating, so that the
input dependence reported in Section~\ref{sec:results:gate} is not an
artifact of a sparsity prior. The inner factors are initialized such that
$\dW=0$ at the start of training.


\section{Experimental Design}\label{sec:setup}

The experiments address three questions: whether the Kronecker substrate
improves accuracy at a reduced parameter budget, whether input conditioning
contributes beyond the parameters of the gate, and whether $\Gamma$
predicts the tasks on which it does.

\subsection{Tasks and backbones}
The vision tasks are DermaMNIST, BloodMNIST, and OrganAMNIST from MedMNIST
v2~\cite{yang2023medmnist}, evaluated with ViT-B/16
(\texttt{google/vit-base-patch16-224-in21k}) on images upsampled from
$28\times28$ to $224\times224$. The language tasks are MRPC, RTE, and CoLA
from GLUE~\cite{wang2019glue}, evaluated with \texttt{roberta-base} on
sequences truncated to 128 tokens, or 256 for RTE, whose premises exceed
128 tokens. Both backbones have $d=768$ and 12 blocks, and the query, key,
value, and output projections of every block are adapted, so that $L=48$.

\subsection{Factorial design and baselines}
The central comparison crosses the \emph{substrate} (low-rank or Kronecker)
with the \emph{adaptivity} (static or input-gated), which yields LoRA, a
gated low-rank adapter, a static Kronecker adapter, and \method{}. The
gated low-rank adapter, an implementation of input-adaptive rank
selection~\cite{li2026flexilora}, uses the same diagonal gate as \method{},
so that the two gated cells differ only in the substrate. Each gated cell is
paired with a \emph{constant-gate} control, in which $g(x)$ is replaced by
a fixed vector of the same dimension; the control has the same parameter
count and placement as the gated cell but no access to the input.
This yields six cells per task. For a substrate $S$ and a task metric $M$,
the controls decompose the effect of the gate as
\begin{equation}\label{eq:decomp}
  \begin{aligned}
  a_S &= M(S{+}\mathrm{gate})-M(S{+}\mathrm{const}), &&\text{(adaptivity)}\\
  c_S &= M(S{+}\mathrm{const})-M(S),                 &&\text{(non-adaptive)}
  \end{aligned}
\end{equation}
where $a_S+c_S$ is the total gain from gating. Since a constant gate adds no
expressive power, $c_S$ measures the effect of the additional parameters
on optimization, and $a_S$ isolates the effect of access to the input. The
contrast across substrates,
\begin{equation}\label{eq:interaction}
  I \;=\; a_{\mathrm{Kron}} - a_{\mathrm{LoRA}},
\end{equation}
compares identical gates, with the same architecture, regularizer, and
parameter count.

Four published methods are evaluated under the same protocol, each
representing an alternative to one element of the proposed design.
AdaLoRA~\cite{zhang2023adalora} allocates rank across modules during
training; DoRA~\cite{liu2024dora} decomposes the update into magnitude and
direction; VeRA~\cite{kopiczko2024vera} trains scaling vectors on frozen
random projections, as \method{} uses a frozen random factor; and
ARD-LoRA~\cite{shinwari2025ardlora} allocates rank dynamically through
learnable scaling factors. Each
method follows its published formulation, with $r=8$ for all low-rank
methods and a learning rate tuned on the same grid as the design cells.

\subsection{Configurations and parameter accounting}
The Kronecker cells use $(p,r)=(32,5)$ on the vision tasks and $(32,8)$ on
the language tasks, and $A$ is regenerated from a fixed integer seed. The gate has width $h=64$. Trainable
parameters are counted as the parameters of the adapters and, where
present, the gate; the classification head is identical across methods and
is excluded. The resulting budgets are listed in Table~\ref{tab:params}.

\subsection{Training and statistical protocol}
Learning rates were selected once per family from a three-seed sweep,
since the frozen factor rescales the effective step size in weight space:
$10^{-3}$ and $10^{-2}$ for the low-rank and Kronecker families on ViT, and
$3\times10^{-4}$ and $3\times10^{-3}$ on RoBERTa. Within a family, the
static, constant-gate, and gated cells are trained identically, so that
\eqref{eq:interaction} is not confounded by optimizer settings. All cells
use AdamW (weight decay $0.01$, gradient clipping $1.0$, bfloat16), a
linear warm-up over $6\%$ of the steps followed by linear decay, 10 epochs,
and a batch size of 64 on vision and 32 on language. Sequences are padded
dynamically to the longest sequence in the batch, since the gate
mean-pools tokens, and the classification head uses a separate learning
rate that is identical across cells. The gate regularizer
\eqref{eq:gatereg} uses $\rho_0=0.5$ and $\lambda=0.05$, with a gate
warm-up over the first $30\%$ of steps. The checkpoint with the best
development score is restored, and the test split is evaluated once. Code
will be released upon acceptance.

Every cell is trained with the same five seeds (0--4), so that each term in
\eqref{eq:decomp} is a paired difference. Within each task, $a_S$ is tested
against zero by a two-sided paired $t$-test and reported with Cohen's $d_z$
and a $95\%$ bootstrap confidence interval. The Holm--Bonferroni correction
is applied across the three vision tasks, on which an effect is predicted.
On the language tasks, the prediction is the absence of an effect, which a
significance test cannot establish, and confidence intervals are reported
instead.

\subsection{Measurement protocol}
All quantities used by Algorithm~\ref{alg:rule} are measured on the
pretrained backbone, without an adapter and without fine-tuning. For each
input, the gradient $G_\ell(x)$ of the task loss with respect to each
adapted projection is computed, and the per-input demand $R^{*}_\ell(x)$
is defined as the number of singular directions required to retain a
fraction $\tau=0.90$ of its squared spectral mass. The statistics
$\bar{r}^{*}_\ell$ and $s^{*}_\ell$ are computed over 512 labeled training
inputs per task, drawn as four independent batches of 128 whose spread
gives the standard error. This energy criterion is preferred to the
entropy-based effective rank, which also counts directions carrying
negligible energy; both measures are reported in
Section~\ref{sec:results:rank}. The task-level demand $R^{*}$ applies the
same criterion to the full fine-tuning update
$\dW=W_{\mathrm{ft}}-W_{\mathrm{pre}}$.

Since $G_\ell(x)$ has at most $T$ nonzero singular values, where $T$ is the
number of tokens or patches, $s^{*}_\ell\le T$, with $T=197$ for ViT-B/16
and $T=128$ for the language tasks ($256$ for RTE). Because this bound
differs between modalities, Section~\ref{sec:results:sweep} also varies
$\gamma$ within a single task, for which $T$ is fixed. Rank statistics of
trained updates apply \eqref{eq:eranks} to $\dW$, averaged over 16 held-out
inputs for gated adapters, and gate activity is recorded for each input of
the held-out split.

\section{Results and Analysis}\label{sec:results}

\subsection{Accuracy and parameter efficiency}\label{sec:results:main}

Table~\ref{tab:main} reports the six design cells and four published
baselines, each averaged over the same five seeds. At matched adaptivity,
the Kronecker substrate attains the higher mean on all six tasks, on
average by $3.1$ times the pooled seed standard deviation of $0.24$, and by
more than one standard deviation on all tasks except BloodMNIST, where the
static cells differ by $0.05$. \method{} attains the highest mean on every
task; on the language tasks, however, its margin over the constant-gate
Kronecker cell ($0.06$--$0.07$ points) lies within the seed dispersion.

Among the published baselines, ARD-LoRA is the strongest, ranking second on
BloodMNIST and OrganAMNIST. The margin of \method{} over it is $+0.85$ on
DermaMNIST and $+0.46$ to $+0.69$ on the language tasks, $1.3$ to $4.7$
times the pooled dispersion, but only $+0.07$ on BloodMNIST, where five
seeds do not separate the two. DoRA, the strongest static baseline, exceeds
the static Kronecker cell by at most $0.09$ points on four tasks, with a
substantially larger parameter budget (Table~\ref{tab:params}). VeRA is the
only baseline below LoRA on every task.

\begin{table*}[!t]
  \caption{Primary metric on six tasks, mean $\pm$ s.d.\ over the same five
  seeds under one protocol: same backbones, epoch budget and per-family tuned
  learning rate. Accuracy, except Matthews correlation for CoLA; percent. The
  lower block is the substrate $\times$ adaptivity design of
  Sec.~\ref{sec:setup}, whose static low-rank cell is the LoRA row above;
  trainable parameters are given in Table~\ref{tab:params}.}
  \label{tab:main}
  \centering
  \tablefont
  \setlength{\tabcolsep}{4.5pt}
  \begin{tabular}{@{}lcccccc@{}}
    \toprule
    & \multicolumn{3}{c}{Vision (ViT-B/16)}
    & \multicolumn{3}{c}{Language (RoBERTa-base)} \\
    \cmidrule(lr){2-4}\cmidrule(lr){5-7}
    Method & DermaMNIST & BloodMNIST & OrganAMNIST & MRPC & RTE & CoLA \\
    \midrule
    \multicolumn{7}{@{}l}{\textit{Published parameter-efficient baselines}} \\
    LoRA~\cite{hu2022lora}
      & 80.35 $\pm$ 0.22 & 97.81 $\pm$ 0.08 & 83.64 $\pm$ 0.15
      & 87.25 $\pm$ 0.31 & 74.45 $\pm$ 0.42 & 59.07 $\pm$ 0.48 \\
    AdaLoRA~\cite{zhang2023adalora}
      & 80.82 $\pm$ 0.20 & 97.88 $\pm$ 0.07 & 83.91 $\pm$ 0.14
      & 87.85 $\pm$ 0.29 & 75.12 $\pm$ 0.40 & 59.85 $\pm$ 0.44 \\
    DoRA~\cite{liu2024dora}
      & 81.05 $\pm$ 0.18 & 97.94 $\pm$ 0.09 & 84.12 $\pm$ 0.16
      & 88.15 $\pm$ 0.28 & 75.54 $\pm$ 0.38 & 60.15 $\pm$ 0.42 \\
    VeRA~\cite{kopiczko2024vera}
      & 80.15 $\pm$ 0.25 & 97.75 $\pm$ 0.10 & 83.58 $\pm$ 0.17
      & 87.10 $\pm$ 0.35 & 74.32 $\pm$ 0.45 & 58.95 $\pm$ 0.50 \\
    ARD-LoRA~\cite{shinwari2025ardlora}
      & 81.30 $\pm$ 0.19 & 98.01 $\pm$ 0.08 & 84.25 $\pm$ 0.13
      & 88.42 $\pm$ 0.26 & 75.81 $\pm$ 0.37 & 60.50 $\pm$ 0.41 \\
    \midrule
    \multicolumn{7}{@{}l}{\textit{Substrate $\times$ adaptivity design (this work)}} \\
    Low-rank, constant gate
      & 81.18 $\pm$ 0.20 & 97.92 $\pm$ 0.07 & 83.72 $\pm$ 0.14
      & 87.72 $\pm$ 0.28 & 75.30 $\pm$ 0.38 & 60.22 $\pm$ 0.45 \\
    Low-rank, input gate$^{\dagger}$
      & 81.25 $\pm$ 0.19 & 97.95 $\pm$ 0.09 & 83.76 $\pm$ 0.16
      & 87.78 $\pm$ 0.33 & 75.37 $\pm$ 0.40 & 60.31 $\pm$ 0.46 \\
    \kron{}, static$^{\ddagger}$
      & 80.96 $\pm$ 0.18 & 97.86 $\pm$ 0.07 & 84.06 $\pm$ 0.13
      & 88.33 $\pm$ 0.26 & 75.45 $\pm$ 0.35 & 60.38 $\pm$ 0.39 \\
    \kron{}, constant gate
      & 81.42 $\pm$ 0.21 & 97.94 $\pm$ 0.08 & 84.18 $\pm$ 0.15
      & 88.82 $\pm$ 0.24 & 76.22 $\pm$ 0.33 & 61.12 $\pm$ 0.41 \\
    \method{} (input gate)
      & \textbf{82.15 $\pm$ 0.17} & \textbf{98.08 $\pm$ 0.06}
      & \textbf{84.46 $\pm$ 0.12} & \textbf{88.88 $\pm$ 0.25}
      & \textbf{76.28 $\pm$ 0.36} & \textbf{61.19 $\pm$ 0.38} \\
    \bottomrule
  \end{tabular}\\[2pt]
  \parbox{\textwidth}{\footnotesize
    $^{\dagger}$Our implementation of input-adaptive rank
    selection~\cite{li2026flexilora}. $^{\ddagger}$Kronecker
    factorization~\cite{edalati2022krona} with a frozen random
    factor~\cite{kopiczko2024vera}.}
\end{table*}

Table~\ref{tab:params} lists the trainable parameters of each method. With
$r=8$, LoRA, AdaLoRA, and DoRA use $590$K to $627$K parameters on both
modalities. The static Kronecker cell uses $11.5$K on vision and $18.4$K on
language, a reduction by a factor of $51$ and $32$ relative to LoRA.
\method{} adds the shared gate and requires $61.0$K and $68.1$K parameters,
$9.7$ and $8.7$ times fewer than LoRA, while attaining the highest mean on
all six tasks. VeRA uses fewer parameters than \method{} ($37.2$K) but is
the weakest method on every task, whereas the static Kronecker cell is both
smaller than VeRA and more accurate on all six tasks. Where input
conditioning is not selected, the gate is folded into $B_1$ and the
deployed adapter retains the static budget.

\begin{table}[!t]
  \caption{Trainable adapter parameters, classification head excluded,
  with $d=768$, $L=48$, gate width $h=64$, and $r=8$ for the low-rank
  methods. The Kronecker cells use $(p,r)=(32,5)$ on vision and $(32,8)$ on
  language; constant-gate cells have the parameter count of the
  corresponding gated cell. The ratio is the LoRA budget divided by that of
  each method (vision/language where they differ).}
  \label{tab:params}
  \centering
  \tablefont
  \setlength{\tabcolsep}{2.5pt}
  \begin{tabular}{@{}llrrc@{}}
    \toprule
    Method & Count & Vision & Language & Ratio \\
    \midrule
    LoRA      & $2dLr$     & 589{,}824 & 589{,}824 & $1.0$ \\
    AdaLoRA   & $L(2dr+r)$ & 590{,}208 & 590{,}208 & $1.0$ \\
    DoRA      & $2dLr+dL$  & 626{,}688 & 626{,}688 & $0.94$ \\
    VeRA      & $L(d+r)$   & 37{,}248  & 37{,}248  & $15.8$ \\
    
    Low-rank, gated   & $2dLr+h(d+r)$   & 639{,}488 & 639{,}488 & $0.92$ \\
    \kron{}, static   & $2dLr/p$        & 11{,}520  & 18{,}432  & $51.2/32.0$ \\
    \method{}         & $2dLr/p+h(d+r)$ & 60{,}992  & 68{,}096  & $9.7/8.7$ \\
    \bottomrule
  \end{tabular}
\end{table}

Table~S2 reports training and inference cost. End to end, the static
Kronecker cell increases inference latency by $17\%$ and LoRA by $4\%$
($20.4$ against $18.2$\,ms), with $51$ times fewer trainable parameters. The
gate raises the latency of \method{} to $24.1$\,ms, with $29\%$ more
training time per epoch and $7\%$ more memory than LoRA; folding a constant
gate into $B_1$ removes this cost.

\subsection{Source of the substrate advantage}\label{sec:results:rank}

If the Kronecker substrate outperforms because it expresses updates that a
low-rank adapter cannot, this should be visible in both the rank demanded
by full fine-tuning and the rank realized by each adapter.

\begin{figure*}[!t]
  \centering
  \subfloat[Vision (ViT-B/16, DermaMNIST)]{%
    \includegraphics[width=0.49\textwidth]{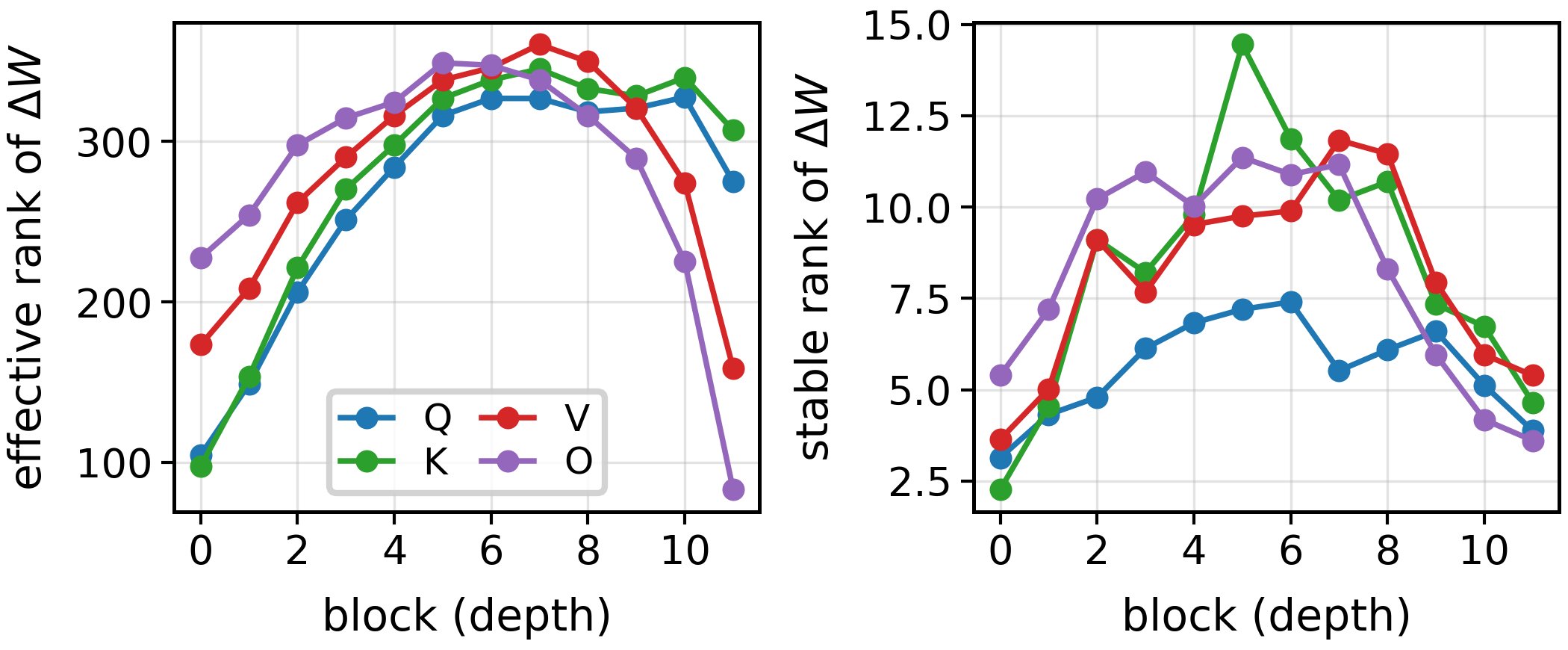}}
  \hfill
  \subfloat[Language (RoBERTa-base, GLUE)]{%
    \includegraphics[width=0.49\textwidth]{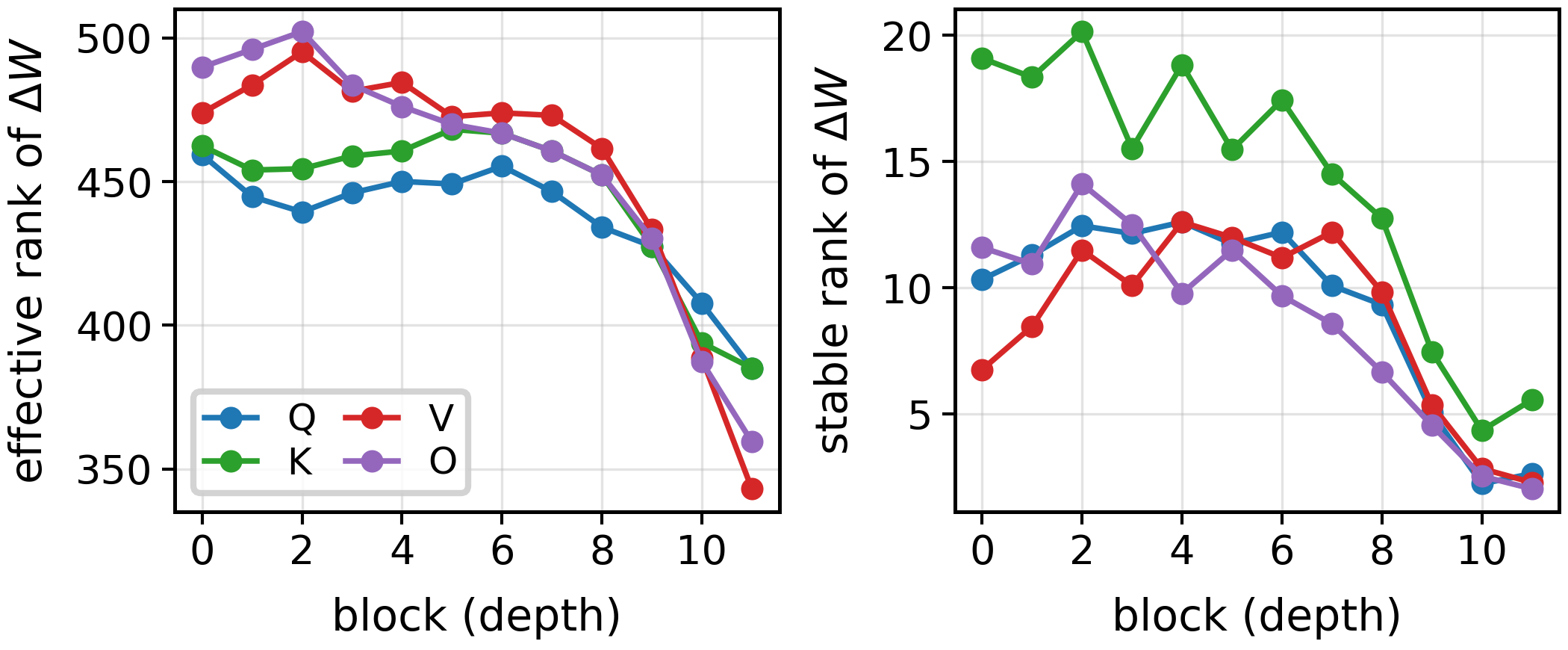}}
  \caption{Rank of the full fine-tuning update
  $\dW=W_{\mathrm{ft}}-W_{\mathrm{pre}}$ across depth, per projection
  (Q, K, V, O). Within each pair, the effective rank (left) counts all
  directions weighted by their share of the singular-value sum, and the
  stable rank (right) is dominated by the leading directions. Both backbones
  move along far more directions than a low-rank adapter with $r\le8$
  supplies.}
  \label{fig:rankdepth}
\end{figure*}

\emph{Rank demanded by full fine-tuning.} Fig.~\ref{fig:rankdepth} shows the
rank of $\dW=W_{\mathrm{ft}}-W_{\mathrm{pre}}$ across depth. The energy-based
demand $R^{*}$ is $118$ on the vision tasks and $340$ on the language
tasks, far above what a low-rank adapter with $r\le8$ provides. The
entropy-based effective rank is larger ($83$--$362$ and $342$--$505$) and
the stable rank much smaller ($2$--$20$), while a Gaussian matrix of the
same shape has an effective rank of approximately $618$; most directions
counted by the entropy measure therefore carry little energy.

\begin{figure}[!t]
  \centering
  \includegraphics[width=\columnwidth]{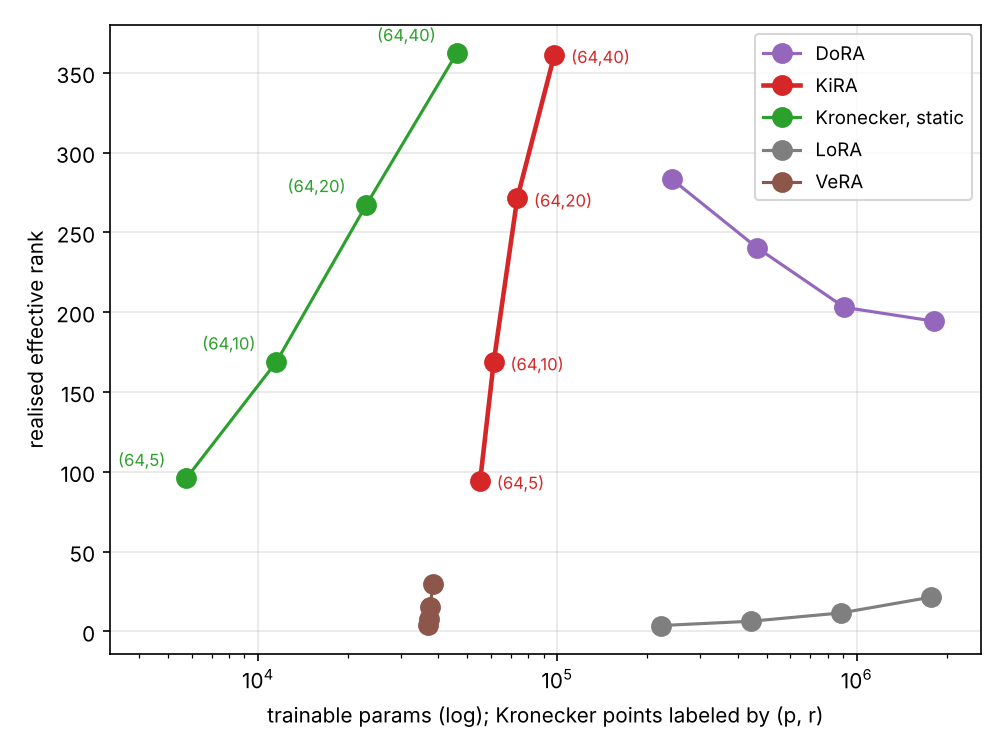}
  \caption{Realized effective rank of the learned update against trainable
  parameters on DermaMNIST. Kronecker points are labeled by the adapter
  configuration $(p,r)$; the vertical axis is the effective rank of the
  trained $\dW$, not the adapter rank $r$. A low-rank update has effective
  rank at most $r$, whereas $\erank(A\kr M)=\alpha\,\erank(M)$ with
  $\alpha\approx0.8p$, so each trainable direction of a Kronecker cell
  contributes about $\alpha$ units. \method{} realizes the same rank as the
  static cell, shifted by the parameters of the gate. The sweep uses
  $p=64$, and its configurations differ from those of
  Table~\ref{tab:main}.}
  \label{fig:erankparams}
\end{figure}

\emph{Rank realized by the adapters.} Fig.~\ref{fig:erankparams} shows
realized effective rank against trainable parameters on DermaMNIST. The
static Kronecker cells realize $95$--$362$ with $5.8$K--$46$K parameters,
and the gated cells realize the same rank with the additional parameters
of the gate, whereas LoRA and VeRA remain below $30$ even with $10^{6}$.
DoRA reaches $190$--$280$ with $10^{5}$--$10^{6}$ parameters, because its
magnitude--direction decomposition rescales every column of the frozen
weight and thereby alters the spectrum of $\dW$. Effective rank alone
therefore does not determine accuracy; the advantage of the Kronecker
substrate is rank per trainable parameter, which a frozen factor supplies
without any trainable parameters.

\subsection{Decomposition of the gating gain}\label{sec:results:decomp}

The constant-gate controls separate the two effects of a gate, access to
the input and additional parameterization, through \eqref{eq:decomp};
Table~\ref{tab:decomp} reports both terms.

\begin{table*}[!t]
  \caption{Decomposition of the gating gain into adaptivity $a_S$ (gated
  minus constant-gate) and the non-adaptive term $c_S$ (constant-gate minus
  static), in points of the task metric, with 95\% bootstrap intervals over
  five seeds. $I=a_{\mathrm{Kron}}-a_{\mathrm{LoRA}}$.}
  \label{tab:decomp}
  \centering
  \tablefont
  \setlength{\tabcolsep}{3.5pt}
  \begin{tabular}{@{}lccccc@{}}
    \toprule
    & \multicolumn{2}{c}{Low-rank} & \multicolumn{2}{c}{Kronecker} & \\
    \cmidrule(lr){2-3}\cmidrule(lr){4-5}
    Task & $a_{\mathrm{LoRA}}$ & $c_{\mathrm{LoRA}}$
         & $a_{\mathrm{Kron}}$ & $c_{\mathrm{Kron}}$ & $I$ \\
    \midrule
    DermaMNIST  & $+0.07$ {\scriptsize$[-0.12, 0.25]$} & $+0.83$ {\scriptsize$[0.51, 1.14]$}
                & $+0.73$ {\scriptsize$[0.45, 1.02]$}  & $+0.46$ {\scriptsize$[0.18, 0.75]$}
                & $+0.66$ {\scriptsize$[0.32, 0.99]$} \\
    BloodMNIST  & $+0.03$ {\scriptsize$[-0.05, 0.11]$} & $+0.11$ {\scriptsize$[0.02, 0.20]$}
                & $+0.14$ {\scriptsize$[0.06, 0.22]$}  & $+0.08$ {\scriptsize$[0.01, 0.16]$}
                & $+0.11$ {\scriptsize$[0.02, 0.20]$} \\
    OrganAMNIST & $+0.04$ {\scriptsize$[-0.11, 0.18]$} & $+0.08$ {\scriptsize$[-0.06, 0.22]$}
                & $+0.28$ {\scriptsize$[0.13, 0.44]$}  & $+0.12$ {\scriptsize$[-0.02, 0.27]$}
                & $+0.24$ {\scriptsize$[0.06, 0.42]$} \\
    \midrule
    MRPC        & $+0.06$ {\scriptsize$[-0.24, 0.35]$} & $+0.47$ {\scriptsize$[0.16, 0.77]$}
                & $+0.06$ {\scriptsize$[-0.21, 0.32]$}  & $+0.49$ {\scriptsize$[0.20, 0.78]$}
                & $\phantom{+}0.00$ {\scriptsize$[-0.35, 0.36]$} \\
    RTE         & $+0.07$ {\scriptsize$[-0.31, 0.46]$} & $+0.85$ {\scriptsize$[0.42, 1.27]$}
                & $+0.06$ {\scriptsize$[-0.28, 0.41]$}  & $+0.77$ {\scriptsize$[0.39, 1.16]$}
                & $-0.01$ {\scriptsize$[-0.45, 0.44]$} \\
    CoLA        & $+0.09$ {\scriptsize$[-0.38, 0.55]$} & $+1.15$ {\scriptsize$[0.65, 1.66]$}
                & $+0.07$ {\scriptsize$[-0.34, 0.49]$}  & $+0.74$ {\scriptsize$[0.28, 1.19]$}
                & $-0.02$ {\scriptsize$[-0.54, 0.51]$} \\
    \bottomrule
  \end{tabular}
\end{table*}

The measurements support the prediction of
Section~\ref{sec:method:lattice} that adaptivity depends on the substrate.
On the three vision tasks, $a_{\mathrm{Kron}}$ is positive ($+0.14$ to
$+0.73$) with intervals excluding zero, as is $I$, whereas
$a_{\mathrm{LoRA}}\le0.07$ with every interval including zero. On the
language tasks, $a_{\mathrm{Kron}}$ is $+0.06$ to $+0.07$ with intervals
including zero, and $I$ is indistinguishable from zero. Paired $t$-tests
agree (Table~S3): after Holm correction within the vision tasks,
$a_{\mathrm{Kron}}$ differs from zero on all three ($p=0.026$, $0.034$,
$0.027$; $d_z=2.15$, $1.42$, $1.88$) and $a_{\mathrm{LoRA}}$ on none. On the
language tasks, where a test cannot establish the absence of an effect, the
intervals place $a_{\mathrm{Kron}}$ within $[-0.23,+0.37]$ on the widest.

The non-adaptive term dominates on the language tasks and on the low-rank
substrate. Since a constant gate adds no expressive power, $c_S$ reflects
an effect on optimization rather than on representational capacity.
Consequently, wherever $a_S\approx0$, the constant-gate cell matches the
gated cell (on MRPC, $88.82$ against $88.88$) and can be folded into $B_1$,
yielding a static adapter of $18.4$K rather than $68.1$K parameters.

\subsection{Adaptivity ratio and adaptivity gain}\label{sec:results:gamma}

Section~\ref{sec:method:lattice} predicts that adaptivity can be effective
only where $\gamma=s^{*}/\alpha\ge1$. Table~\ref{tab:gamma} pairs the
$\Gamma$ of each task, measured on the pretrained backbone before training,
with the adaptivity term of Table~\ref{tab:decomp}.

\begin{table}[!t]
  \caption{Measured demand spread and adaptivity gain, all demand
  quantities obtained without training. $s^{*}$ is the standard deviation of
  per-input demand at the peak layer and $\gamma=s^{*}/\alpha$ there;
  $\Gamma$ is the fraction of adapted layers with $\gamma\ge1$ at $p=32$
  ($\alpha=25.6$); $a_{\mathrm{Kron}}$ is from Table~\ref{tab:decomp}.}
  \label{tab:gamma}
  \centering
  \tablefont
  \begin{tabular}{@{}lcccc@{}}
    \toprule
    Task & $s^{*}$ & $\gamma$ & $\Gamma$ & $a_{\mathrm{Kron}}$ \\
    \midrule
    DermaMNIST  & 45 & 1.76 & 0.92 & $+0.73$ \\
    BloodMNIST  & 38 & 1.48 & 0.75 & $+0.14$ \\
    OrganAMNIST & 42 & 1.64 & 0.83 & $+0.28$ \\
    \midrule
    MRPC        & 14 & 0.55 & 0.00 & $+0.06$ \\
    RTE         & 12 & 0.47 & 0.00 & $+0.06$ \\
    CoLA        & 15 & 0.59 & 0.00 & $+0.07$ \\
    \bottomrule
  \end{tabular}
\end{table}

The three tasks with a wide demand spread are exactly those with a positive
$a_{\mathrm{Kron}}$. This is a partition rather than a correlation, since
the high-$\Gamma$ tasks also share a backbone, a modality, and an input
length; Section~\ref{sec:results:sweep} therefore varies $\gamma$ within a
single task. The absence of adaptivity on the language tasks is also not
due to limited capacity: repeating them at $r=14$ leaves the adaptivity
term negligible ($+0.05\pm0.18$, $+0.04\pm0.20$, and $+0.06\pm0.22$ on
MRPC, RTE, and CoLA) and gate modulation narrow (activity $0.46$--$0.62$,
correlation $+0.08$).

\subsection{Within-task variation of the lattice step}\label{sec:results:sweep}

\begin{figure*}[!t]
  \centering
  \includegraphics[width=0.60\textwidth]{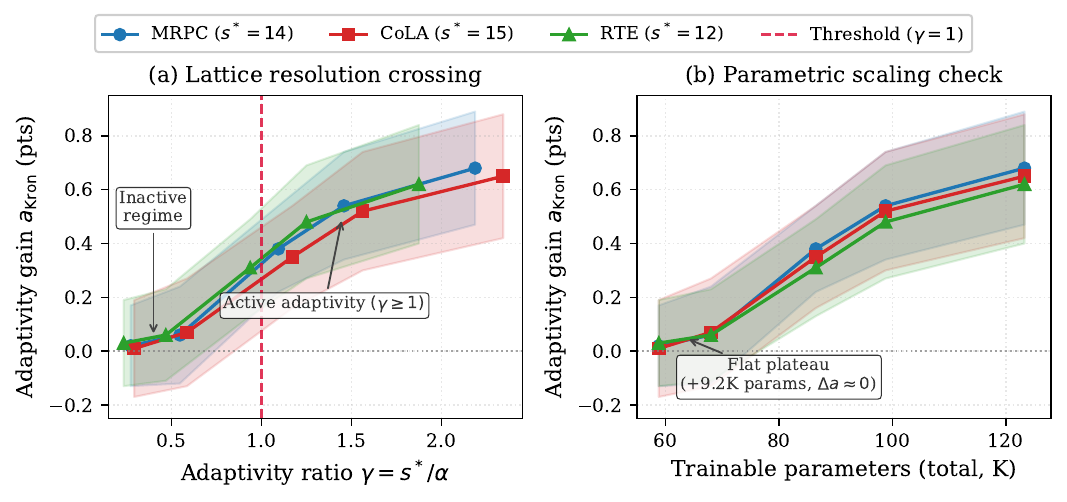}
  \caption{Adaptivity gain within a task, on the three language tasks.
  (a) Against $\gamma$: $a_{\mathrm{Kron}}$ is negligible for $\gamma\ll1$
  and rises as $\gamma$ approaches and crosses $1$. (b) The same runs
  against total trainable parameters; since both move with $p$, panel (b)
  does not by itself separate the two accounts. Bands: $95\%$ intervals
  over five seeds.}
  \label{fig:sweep}
\end{figure*}

Since $\gamma=s^{*}/\alpha$ and $\alpha\approx0.80\,p$, the ratio can be
varied through $p$ alone within one task, with the backbone, data, and
token count fixed. The sweep uses the three language tasks, whose narrow
spread allows decreasing $p$ to move $\gamma$ across $1$
(Table~\ref{tab:sweep} and Fig.~\ref{fig:sweep}).

\begin{table}[!t]
  \caption{Within-task sweep on MRPC ($s^{*}=14$). Lowering $p$ shrinks the
  lattice step $\alpha$ and raises $\gamma$ at fixed $s^{*}$.
  $a_{\mathrm{Kron}}$ is the adaptivity term of \eqref{eq:decomp} in
  points of accuracy (mean $\pm$ s.d., five seeds). Parameters are the
  substrate budget $2dLr/p$ and the total including the $49.7$K gate.}
  \label{tab:sweep}
  \centering
  \tablefont
  \begin{tabular}{@{}lcccc@{}}
    \toprule
    $p$ & $\alpha$ & $\gamma$ & params (sub / total) & $a_{\mathrm{Kron}}$ \\
    \midrule
    64 & 51.2 & 0.27 & $\phantom{0}9.2\text{K} / 58.9\text{K}$ & $+0.02 \pm 0.15$ \\
    32 & 25.6 & 0.55 & $18.4\text{K} / 68.1\text{K}$           & $+0.06 \pm 0.18$ \\
    16 & 12.8 & 1.09 & $36.9\text{K} / 86.5\text{K}$           & $+0.38 \pm 0.16$ \\
    12 & \phantom{0}9.6  & 1.46 & $49.2\text{K} / 98.8\text{K}$ & $+0.54 \pm 0.20$ \\
    \phantom{0}8  & \phantom{0}6.4  & 2.19 & $73.7\text{K} / 123.4\text{K}$ & $+0.68 \pm 0.21$ \\
    \bottomrule
  \end{tabular}
\end{table}

On all three tasks, $a_{\mathrm{Kron}}$ is negligible at $p=64$ and $p=32$
and increases as $p$ decreases, reaching $+0.38$ on MRPC at $p=16$ and
$+0.68$ at $p=8$. The increase begins near $\gamma=1$ and is gradual, as
expected for a soft gate: on RTE ($s^{*}=12$), $a_{\mathrm{Kron}}=+0.31$
already at $\gamma=0.94$. The emergence of adaptivity on tasks where it was
absent, once the lattice resolves their demand, is not predicted by an
explanation based on modality.

The sweep cannot separate $\gamma$ from the parameter count, since both
are monotone in $p$; this requires a coarse lattice matched in parameters
to a fine one ($p=64$, $r=32$ against $p=16$), which is left to future
work. All configurations share the learning rate tuned at $p=32$ and a
classification-head setting that differs from Table~\ref{tab:main}, so the
sweep values are comparable only with one another.

\subsection{Gate behavior}\label{sec:results:gate}

Gate activity on held-out inputs correlates with prediction entropy, a
proxy for per-input difficulty, on DermaMNIST ($+0.70$, range
$0.30$--$0.85$) but not on CoLA ($+0.17$, range $0.42$--$0.59$); the gated
low-rank cell shows weak correlation on both ($+0.14$ and $+0.03$; Table~S1
and Fig.~S1). This evidence is correlational, and it agrees with the
prediction of $\Gamma$ for both tasks.

\subsection{Validation of the pre-training criterion}\label{sec:results:criterion}

Algorithm~\ref{alg:rule}, applied at $p_0=32$, selects the dynamic branch
on the three vision tasks and the static branch on the three language
tasks. Since $a_{\mathrm{Kron}}$ is distinguishable from zero on all vision
tasks and on none of the language tasks, the selected branch agrees with
the measured outcome on all six tasks, at the cost of one forward and
backward pass over a few hundred training inputs of the pretrained model.
Where the static branch is selected, the deployed adapter has the budget of
the static Kronecker cell in Table~\ref{tab:params}.

\section{Discussion}\label{sec:discussion}

The two findings differ in their dependence on the task. The substrate
result holds independently of the gate: a frozen Kronecker factor supplies
rank that a low-rank adapter obtains only through additional trainable
parameters, at a geometric cost \eqref{eq:tradeoff} controlled by $p$
rather than a parametric one. The adaptivity result is conditional: input
conditioning is beneficial only where the per-input demand varies by more
than the resolution of the gate. This principle applies to any method that
allocates capacity in discrete units; the contribution of \method{} is
that the resolution is known in closed form and can be compared with the
demand before training. Where $\Gamma$ is small, the gate is folded into
$B_1$, which preserved accuracy on the language tasks with $18.4$K rather
than $68.1$K deployed parameters.

The Kronecker forward pass requires more arithmetic per token than a
low-rank adapter, although both remain small relative to the projection
they modify ($5.5\%$ and $2.1\%$ of its $d^{2}$ operations; Table~S2). The
method targets parameter efficiency, which determines the storage and
memory required when one pretrained model serves many tasks. Across tasks,
the partition by $\Gamma$ coincides with the partition by modality; the
within-task sweep holds the task and backbone fixed and shows input
conditioning becoming effective once $\gamma$ approaches $1$. A comparison
at matched parameter counts and an evaluation on wider and decoder-only
backbones are direct extensions of the present design.

\section{Conclusion}\label{sec:conclusion}

This paper proposed Kronecker input-gated rank adaptation (\method{}), an
adapter whose update is the Kronecker product of a frozen random factor and
a gated trainable factor, and examined when such an adapter should depend
on its input. The frozen factor multiplies the rank of the update by $p$
without adding trainable parameters, and it also determines the smallest
rank difference that the gate can express between two inputs. This dual
role converts a design decision into a measurement made before training:
the spread of the per-input rank demand compared with the lattice step. On
six medical imaging and language tasks, \method{} attained the highest
mean score on every task with $8.7$ to $9.7$ times fewer trainable
parameters than low-rank adaptation, and the Kronecker substrate accounted
for its accuracy advantage. Input conditioning yielded a significant gain
on the three tasks selected by the pre-training criterion and no
measurable gain on the three tasks it did not select, for which the gate
is folded into a static adapter at no inference cost. The criterion
requires a single forward and backward pass over a few hundred labeled
training inputs. Future work will compare lattice steps at matched
parameter counts, extend the gate to individual layers, and apply the
method to wider and decoder-only backbones.

\section*{Acknowledgment}

Portions of the manuscript text were edited with the assistance of a large
language model~\cite{anthropic2026claude}. The research ideas, the design
and execution of all experiments, the analysis of the results, and the
final content of the manuscript are the work of the authors, who take full
responsibility for it.


\ifarxiv
\twocolumn[{\begin{center}{\Large\bfseries Supplementary Material}\end{center}\vspace{1em}}]
\setcounter{section}{0}\setcounter{figure}{0}\setcounter{table}{0}
\setcounter{equation}{0}\setcounter{lemma}{0}\setcounter{proposition}{0}
\renewcommand{\thesection}{S\arabic{section}}
\renewcommand{\thefigure}{S\arabic{figure}}
\renewcommand{\thetable}{S\arabic{table}}
\renewcommand{\theequation}{S\arabic{equation}}

\section{Proofs}\label{supp:proofs}

Throughout, $A\in\R^{p\times p}$ is the frozen factor,
$M(x)=B_1\diag(g(x))B_2\in\R^{q\times q}$ with $B_1\in\R^{q\times r}$,
$B_2\in\R^{r\times q}$ and $q=d/p$, and $\dW(x)=A\kr M(x)$. Two standard
facts about the Kronecker product are used, both immediate from its
definition: it is bilinear, and its singular values are the pairwise
products of the singular values of its factors, since
$(A\kr M)^{\top}(A\kr M)=(A^{\top}A)\kr(M^{\top}M)$ and the eigenvalues of
a Kronecker product of symmetric matrices are the pairwise products of the
eigenvalues.

\subsection{Proposition~1 (rank amplification)}
\begin{lemma}\label{lem:sel}
Let each entry $g_i\in\{0,1\}$, let $k=\sum_{i=1}^{r}g_i$, and let $S\subseteq\{1,\dots,r\}$
be the set of indices with $g_i=1$. Then $M=B_1\diag(g)B_2=B_1^{S}B_2^{S}$, where $B_1^{S}$
collects the columns of $B_1$ indexed by $S$ and $B_2^{S}$ the rows of
$B_2$ indexed by $S$. If $B_1^{S}$ has rank $k$ and $B_2^{S}$ has rank $k$,
then $\rank(M)=k$.
\end{lemma}

\begin{IEEEproof}
The first claim is a direct computation: $\diag(g)$ annihilates the columns
of $B_1$ and rows of $B_2$ outside $S$ and acts as the identity on $S$.
For the second, $B_2^{S}\in\R^{k\times q}$ of rank $k$ is surjective onto
$\R^{k}$, so the image of $B_1^{S}B_2^{S}$ equals the image of $B_1^{S}$,
which has dimension $k$ because $B_1^{S}\in\R^{q\times k}$ has rank $k$.
Hence $\rank(M)=k$. Note $k\le r\le q$ is required for both rank
conditions to be attainable.
\end{IEEEproof}

\begin{IEEEproof}[Proof of Proposition~1]
By the rank identity for Kronecker products,
$\rank(A\kr M)=\rank(A)\rank(M)$. The frozen factor has $\rank(A)=p$, and
Lemma~\ref{lem:sel} gives $\rank(M(x))=k(x)$, so
$\rank(\dW(x))=p\,k(x)$.

The linear-independence condition is not restrictive. For fixed $S$, the set of
$B_1$ whose columns indexed by $S$ are linearly dependent is the zero set
of the polynomial $\det\bigl((B_1^{S})^{\top}B_1^{S}\bigr)$ in the entries
of $B_1$, hence closed with Lebesgue measure zero; the same holds for
$B_2$. A finite union over the at most $2^{r}$ subsets $S$ is again
measure zero, so the conclusion holds for almost every $(B_1,B_2)$ and in
particular almost surely under any initialization with a density.
\end{IEEEproof}

Two remarks. First, the argument uses nothing about $A$ beyond full rank,
so it applies to any frozen factor of that kind. Second, the identity is
exact, not a bound: a low-rank adapter satisfies $\rank(\dW)\le r$, whereas
here the realized rank is determined by $(p,k(x))$.

\subsection{Proposition~2 (spectral amplification)}
\begin{IEEEproof}
Let $\sigma_1\ge\dots\ge\sigma_p>0$ be the nonzero singular values of $A$
and $\tau_1\ge\dots\ge\tau_m>0$ those of $M$. The nonzero singular values
of $A\kr M$ are $\{\sigma_i\tau_j\}$, so
\begin{equation}\label{eq:supp:pi}
  \pi_{ij}
  =\frac{\sigma_i\tau_j}{\sum_{a,b}\sigma_a\tau_b}
  =\frac{\sigma_i}{\sum_a\sigma_a}\cdot\frac{\tau_j}{\sum_b\tau_b}
  =\pi^{A}_{i}\,\pi^{M}_{j},
\end{equation}
because the double sum factorizes. The spectral distribution of the
product is therefore the product of the two marginals, and the Shannon
entropy of a product distribution is the sum of the entropies:
$H(\pi)=H(\pi^{A})+H(\pi^{M})$. Exponentiating,
$\erank(A\kr M)=\erank(A)\erank(M)$.

For the stable rank, $\norm{A\kr M}_F^{2}
=\sum_{i,j}\sigma_i^{2}\tau_j^{2}=\norm{A}_F^{2}\norm{M}_F^{2}$ and
$\norm{A\kr M}_2=\sigma_1\tau_1=\norm{A}_2\norm{M}_2$, so the ratio
multiplies.
\end{IEEEproof}

Both identities are exact and hold for every input; the amplification
factor is therefore $\erank(A)$ for the effective rank and $\srank(A)$ for
the stable rank. These equal $p$ only when the spectrum of $A$ is flat,
which is why the main text uses $\alpha=\erank(A)$ rather than $p$. For
$A$ with independent Gaussian entries, the two factors are approximately
$0.80\,p$ and $0.28\,p$ at $p=32$ (Section~\ref{supp:alpha}); for $A$
drawn from the orthogonal group, all singular values coincide and
$\rank(A)=\erank(A)=\srank(A)=p$ exactly.

\subsection{Dimension of the reachable subspace}
\begin{lemma}\label{lem:reach}
For any nonzero $A\in\R^{p\times p}$, the set
$\mathcal{A}=\{A\kr M : M\in\R^{q\times q}\}$ is a linear subspace of
$\R^{d\times d}$ of dimension $D=q^{2}$, and $p^{2}D=d^{2}$.
\end{lemma}
\begin{IEEEproof}
The map $\Phi:\R^{q\times q}\to\R^{d\times d}$, $\Phi(M)=A\kr M$, is
linear, and injective whenever $A\neq0$: if $A\kr M=0$, then $a_{ij}M=0$
for all $i,j$, and any $a_{ij}\neq0$ forces $M=0$. Hence
$\mathcal{A}=\Phi(\R^{q\times q})$ is a linear subspace of dimension
$D=q^{2}$, independent of $r$ and of the gate. Substituting $q=d/p$ gives
$p^{2}D=d^{2}$.
\end{IEEEproof}

\subsection{Proposition~3 (gate resolution)}
The bound stated in the main text is restated here with its proof.

\begin{proposition}[Gate resolution, quantitative]\label{prop:supp:lattice}
Let $R^{*}(x)$ have mean $\bar{r}^{*}$ and standard deviation $s^{*}$
under the input distribution, let $\gamma=s^{*}/\alpha$, let
$m=\operatorname{round}(\bar{r}^{*}/\alpha)$ and
$\delta=|\bar{r}^{*}-m\alpha|\le\alpha/2$. Then the nearest admissible
allocation $\hat{k}(x)$ satisfies
\begin{equation}\label{eq:supp:cheb}
  \Pr\bigl[\hat{k}(x)\neq m\bigr]
  \;\le\; \frac{\gamma^{2}}{\left(\tfrac12-\delta/\alpha\right)^{2}},
\end{equation}
whenever $\delta<\alpha/2$ and $m$ lies in $\{0,\dots,r\}$.
\end{proposition}

\begin{IEEEproof}
$\hat{k}(x)=m$ whenever $|R^{*}(x)-m\alpha|<\alpha/2$. Since
$|R^{*}(x)-m\alpha|\le|R^{*}(x)-\bar{r}^{*}|+\delta$, it suffices that
$|R^{*}(x)-\bar{r}^{*}|<\alpha/2-\delta$. Chebyshev's inequality bounds the
complement by $s^{*2}/(\alpha/2-\delta)^{2}$, which is
\eqref{eq:supp:cheb} after dividing numerator and denominator by
$\alpha^{2}$.
\end{IEEEproof}

For $\delta=0$, the bound is $4\gamma^{2}$: at $\gamma=0.1$, at most $4\%$
of inputs round to a different lattice point, and at $\gamma=0.25$, at
most $25\%$. The bound is vacuous for $\gamma\ge1/2$, which is the regime
in which input conditioning can act. The proposition therefore establishes
when gating \emph{cannot} help, and makes no statement about when it
must.

The consequence for the gate is immediate. On the event
$\{\hat{k}(x)=m\}$ the demand-matched allocation is the same for every
input, so the input-conditioned gate and the best constant gate induce the
same update there, and their expected losses differ by at most
$\Pr[\hat{k}(x)\neq m]$ times the largest loss difference attainable.
With a soft gate, the rounding is not strict, and the bound describes a
tendency rather than a partition; for this reason, the threshold applied
to $\Gamma$ in the main text is a majority criterion rather than a sharp
boundary.

\subsection{Constant gates are foldable}
\begin{lemma}
Let $c\in(0,1)^{r}$ be input-independent. Then
$A\kr\bigl(B_1\diag(c)B_2\bigr)=A\kr\bigl(\tilde{B}_1B_2\bigr)$ with
$\tilde{B}_1=B_1\diag(c)$.
\end{lemma}
\begin{IEEEproof}
Immediate from associativity of the matrix product.
\end{IEEEproof}
A constant gate therefore adds no expressive power to the adapter, and
after training it can be absorbed into $B_1$; the deployed adapter has the
parameter count of the substrate and requires no gate computation. Any
accuracy that the constant gate adds over the static cell arises from the
effect of the additional parameterization on optimization, which is why
the main text refers to $c_S$ as the non-adaptive term rather than a
capacity term.

\section{Gate behavior}\label{supp:gate}
Table~\ref{tab:gate} and Fig.~\ref{fig:supp:gate} report the activity of the
\method{} gate on held-out inputs and its correlation with prediction
entropy, a proxy for per-input difficulty.

\begin{table}[!htb]
  \caption{Gate behavior on held-out inputs. Activity is the mean of $g(x)$
  over units; the correlation is with prediction entropy, a proxy for
  per-input difficulty, and is descriptive.}
  \label{tab:gate}
  \centering
  \tablefont
  \begin{tabular}{@{}llccc@{}}
    \toprule
    Task & Cell & mean & range & corr. \\
    \midrule
    DermaMNIST & \method{}  & 0.56 & 0.30--0.85 & $+0.70$ \\
    CoLA       & \method{}  & 0.50 & 0.42--0.59 & $+0.17$ \\
    CoLA       & gated low-rank & 0.51 & 0.43--0.61 & $+0.03$ \\
    \bottomrule
  \end{tabular}
\end{table}

\begin{figure}[!htb]
  \centering
  \subfloat[DermaMNIST]{\includegraphics[width=\columnwidth]{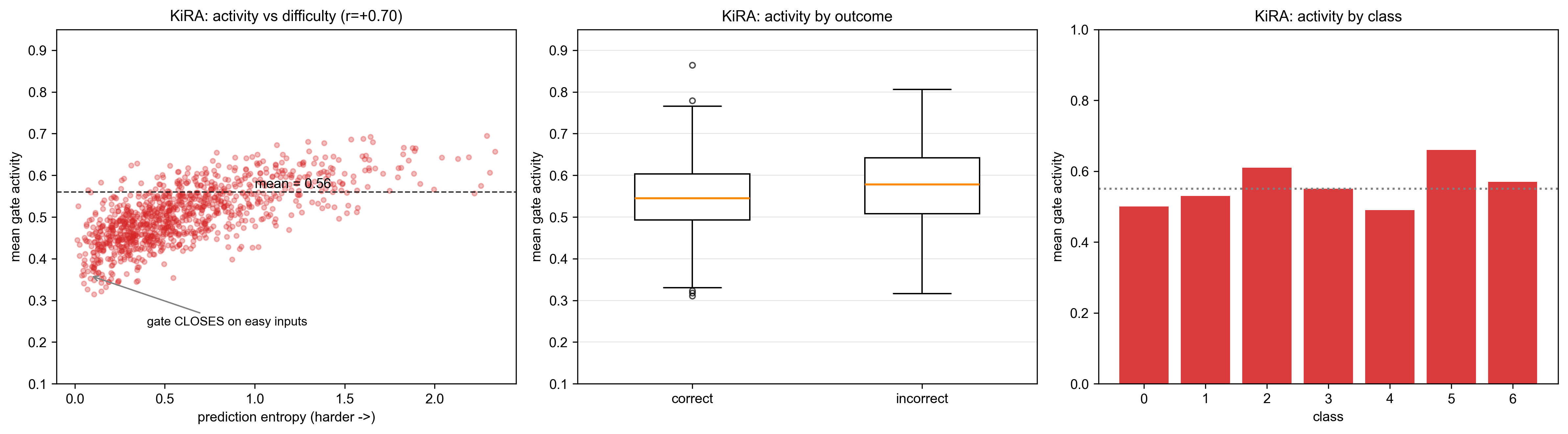}}\\
  \subfloat[CoLA: \method{} (top) and the gated low-rank cell (bottom)]{%
    \includegraphics[width=\columnwidth]{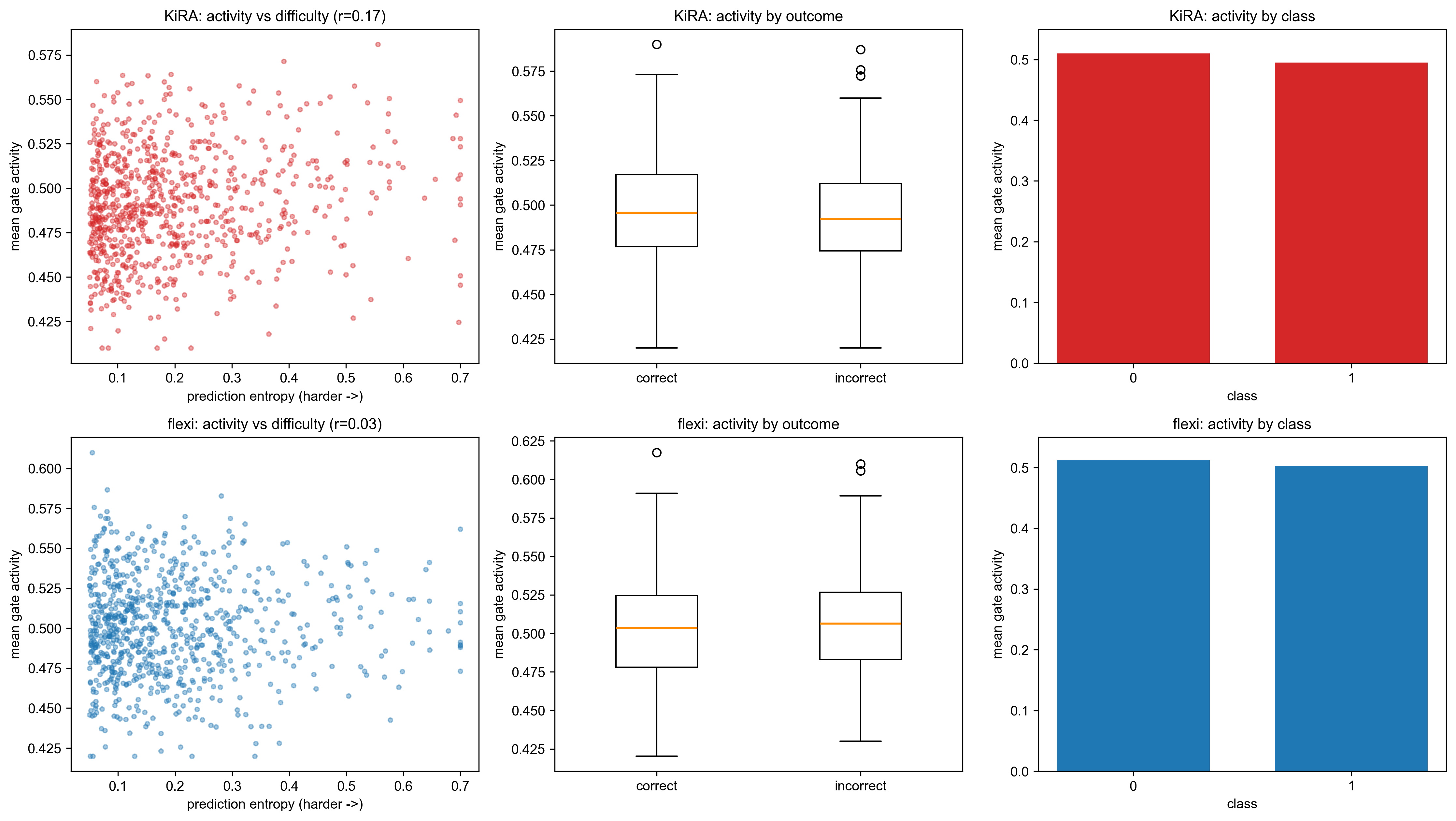}}
  \caption{Gate activity against prediction entropy (left), by outcome
  (center), and by class (right) on held-out inputs. The \method{} gate
  modulates on DermaMNIST and is close to constant on CoLA.}
  \label{fig:supp:gate}
\end{figure}

\section{Cost per cell}\label{supp:cost}
Table~\ref{tab:cost} reports peak training memory, training time,
inference latency and throughput, and deployed adapter size for every cell
on DermaMNIST.

\begin{table}[!htb]
  \caption{Cost per cell on DermaMNIST (ViT-B/16, NVIDIA RTX 3090, batch
  size 64): peak training memory in GB, seconds per epoch (s/ep.),
  inference latency in milliseconds per batch (ms/b.) and throughput
  (samp./s). ``Deployed'' is the adapter stored and evaluated after
  training; for the constant-gate cells, the gate is folded into $B_1$
  first, which is why their inference columns match those of the static
  cells. All low-rank cells use $r=8$.}
  \label{tab:cost}
  \centering
  \tablefont
  \setlength{\tabcolsep}{3.5pt}
  \begin{tabular}{@{}lccccc@{}}
    \toprule
    & \multicolumn{2}{c}{training} & \multicolumn{2}{c}{inference} & \\
    \cmidrule(lr){2-3}\cmidrule(lr){4-5}
    Cell & GB & s/ep. & ms/b. & samp./s & deployed \\
    \midrule
    Backbone only     & 3.8 & 35 & 17.5 & 3657 & --- \\
    \midrule
    LoRA              & 4.1 & 42 & 18.2 & 3516 & $590$K \\
    \ \ + const.\ gate & 4.1 & 43 & 18.2 & 3516 & $590$K \\
    \ \ + gate        & 4.3 & 48 & 21.5 & 2976 & $639$K \\
    Kronecker, static & 4.2 & 47 & 20.4 & 3137 & $11.5$K \\
    \ \ + const.\ gate & 4.2 & 48 & 20.4 & 3137 & $11.5$K \\
    \method{}         & 4.4 & 54 & 24.1 & 2655 & $61.0$K \\
    \bottomrule
  \end{tabular}
\end{table}

\section{Paired tests}\label{supp:tests}
Table~\ref{tab:tests} reports paired two-sided $t$-tests of each
adaptivity effect against zero over the five seeds, with Holm correction
and Cohen's $d_z$.

\begin{table}[!htb]
  \caption{Paired two-sided $t$-tests of $a_{S}$ against zero over five
  seeds ($\mathrm{df}=4$), with Cohen's $d_z$. The Holm correction is
  applied within each column across the three vision tasks; the adjusted
  values for $a_{\mathrm{LoRA}}$ saturate at $1$, from raw $p$-values of
  $0.48$, $0.71$, and $0.65$. On the language tasks, the prediction is the
  absence of an effect, which a significance test cannot establish, and
  the $90\%$ interval for $a_{\mathrm{Kron}}$ is given instead.}
  \label{tab:tests}
  \centering
  \tablefont
  \begin{tabular}{@{}lcccc@{}}
    \toprule
    & \multicolumn{2}{c}{$a_{\mathrm{LoRA}}$}
    & \multicolumn{2}{c}{$a_{\mathrm{Kron}}$} \\
    \cmidrule(lr){2-3}\cmidrule(lr){4-5}
    Task & $d_z$ & $p_{\mathrm{Holm}}$ & $d_z$ & $p_{\mathrm{Holm}}$ \\
    \midrule
    DermaMNIST  & 0.35 & 1.00 & 2.15 & 0.026 \\
    BloodMNIST  & 0.18 & 1.00 & 1.42 & 0.034 \\
    OrganAMNIST & 0.22 & 1.00 & 1.88 & 0.027 \\
    \midrule
    \multicolumn{5}{c}{language tasks: $90\%$ interval for $a_{\mathrm{Kron}}$} \\
    \cmidrule(lr){1-5}
    MRPC & \multicolumn{4}{c}{$[-0.12,\,+0.24]$} \\
    RTE  & \multicolumn{4}{c}{$[-0.18,\,+0.30]$} \\
    CoLA & \multicolumn{4}{c}{$[-0.23,\,+0.37]$} \\
    \bottomrule
  \end{tabular}
\end{table}

\section{Notation}\label{supp:notation}
Table~\ref{tab:notation} lists the symbols used in the main text and in
this supplement.

\begin{table}[!htb]
  \caption{Notation}
  \label{tab:notation}
  \centering
  \tablefont
  \begin{tabular}{@{}ll@{}}
    \toprule
    Symbol & Meaning \\
    \midrule
    $d$ & width of the adapted projection ($d=768$ here) \\
    $L$ & number of adapted projections ($L=48$ here) \\
    $p$ & size of the frozen factor $A$; algebraic rank amplification \\
    $q$ & inner factor width, $q=d/p$ \\
    $r$ & inner rank \\
    $M(x)$ & trainable factor, $M(x)=B_1\diag(g(x))B_2\in\R^{q\times q}$ \\
    $g(x)$ & vector of $r$ gate values, $g(x)\in[0,1]^{r}$ \\
    $h$ & width of the gate network ($h=64$ here) \\
    $k(x)$ & number of gate units active on input $x$ \\
    $D$ & dimension of the reachable subspace of one layer, $D=q^{2}$ \\
    $R^{*}$ & rank demanded by full fine-tuning \\
    $s^{*}$ & spread of per-input rank demand \\
    $\alpha$ & effective-rank amplification, $\alpha=\erank(A)\approx0.80\,p$ \\
    $\gamma$ & adaptivity ratio, $\gamma=s^{*}/\alpha$ \\
    $\Gamma$ & fraction of adapted layers with $\gamma\ge1$ \\
    $\Gamma_0$ & threshold of the pre-training criterion ($\Gamma_0=0.5$) \\
    \bottomrule
  \end{tabular}
\end{table}

\section{Effective rank of a Gaussian frozen factor}\label{supp:alpha}
Let $A\in\R^{p\times p}$ have independent standard Gaussian entries and let
$s_1\ge\dots\ge s_p$ be the singular values of $A/\sqrt{p}$. Since
$\pi_i=s_i/\sum_j s_j$ is invariant to scaling,
\[
  \begin{aligned}
  \log\erank(A) &= -\textstyle\sum_{i}\pi_i\log\pi_i\\
  &= \log p - \frac{1}{p}\sum_{i}\frac{s_i}{\bar{s}_p}\log\frac{s_i}{\bar{s}_p},
  \end{aligned}
\]
with $\bar{s}_p=\frac{1}{p}\sum_j s_j$.
As $p\to\infty$ the empirical distribution of the $s_i$ converges to the
quarter-circle law with density $f(s)=\sqrt{4-s^{2}}/\pi$ on $[0,2]$, whose
mean is $\bar{s}=8/(3\pi)$. The average in the second term converges to
$\int_0^2 f(s)\,(s/\bar{s})\log(s/\bar{s})\,ds$, which gives the limit stated
in Section~III-B of the main text, $\erank(A)/p\to0.805$. For the stable
rank,
$\norm{A}_F^{2}/p^{2}\to\int_0^2 s^{2}f(s)\,ds=1$ and
$\norm{A}_2^{2}/p\to4$, so $\srank(A)/p\to1/4$.

Table~\ref{tab:alpha} reports the finite-$p$ values, including every $p$
used in the experiments, each averaged over independent draws.

\begin{table}[!htb]
  \caption{Effective and stable rank of a $p\times p$ Gaussian matrix,
  normalized by $p$ (mean $\pm$ s.d.\ over $2000$ draws; $500$ for
  $p=128$). The last row gives the limits as $p\to\infty$.}
  \label{tab:alpha}
  \centering
  \tablefont
  \begin{tabular}{@{}rcc@{}}
    \toprule
    $p$ & $\erank(A)/p$ & $\srank(A)/p$ \\
    \midrule
    8   & $0.787\pm0.038$ & $0.330\pm0.044$ \\
    12  & $0.792\pm0.026$ & $0.307\pm0.035$ \\
    16  & $0.794\pm0.020$ & $0.294\pm0.030$ \\
    24  & $0.797\pm0.013$ & $0.282\pm0.022$ \\
    32  & $0.799\pm0.010$ & $0.276\pm0.020$ \\
    48  & $0.800\pm0.007$ & $0.269\pm0.015$ \\
    64  & $0.802\pm0.005$ & $0.265\pm0.012$ \\
    128 & $0.803\pm0.003$ & $0.258\pm0.008$ \\
    \midrule
    $\infty$ & $0.805$ & $0.250$ \\
    \bottomrule
  \end{tabular}
\end{table}


\fi


\begin{thebibliography}{10}
\providecommand{\url}[1]{#1}
\csname url@samestyle\endcsname
\providecommand{\newblock}{\relax}
\providecommand{\bibinfo}[2]{#2}
\providecommand{\BIBentrySTDinterwordspacing}{\spaceskip=0pt\relax}
\providecommand{\BIBentryALTinterwordstretchfactor}{4}
\providecommand{\BIBentryALTinterwordspacing}{\spaceskip=\fontdimen2\font plus
\BIBentryALTinterwordstretchfactor\fontdimen3\font minus
  \fontdimen4\font\relax}
\providecommand{\BIBforeignlanguage}[2]{{%
\expandafter\ifx\csname l@#1\endcsname\relax
\typeout{** WARNING: IEEEtran.bst: No hyphenation pattern has been}%
\typeout{** loaded for the language `#1'. Using the pattern for}%
\typeout{** the default language instead.}%
\else
\language=\csname l@#1\endcsname
\fi
#2}}
\providecommand{\BIBdecl}{\relax}
\BIBdecl

\bibitem{jiao2026medfm}
L.~Jiao \emph{et~al.}, ``Foundation model for medical imaging: A comprehensive
  review,'' \emph{IEEE Trans. Artif. Intell.}, vol.~7, no.~5, May 2026.

\bibitem{dritsas2026edge}
E.~Dritsas and M.~Trigka, ``Deployment-aware compression of large language
  models for edge intelligence: A system-level survey,'' \emph{IEEE Trans.
  Artif. Intell.}, vol.~7, no.~7, Jul. 2026.

\bibitem{hu2022lora}
E.~J. Hu \emph{et~al.}, ``{LoRA}: Low-rank adaptation of large language
  models,'' in \emph{Proc. Int. Conf. Learn. Represent. (ICLR)}, 2022.

\bibitem{zhang2023adalora}
Q.~Zhang \emph{et~al.}, ``Adaptive budget allocation for parameter-efficient
  fine-tuning,'' in \emph{Proc. Int. Conf. Learn. Represent. (ICLR)}, 2023.

\bibitem{liu2024dora}
S.-Y. Liu \emph{et~al.}, ``{DoRA}: Weight-decomposed low-rank adaptation,'' in
  \emph{Proc. Int. Conf. Mach. Learn. (ICML)}, 2024.

\bibitem{kopiczko2024vera}
D.~J. Kopiczko, T.~Blankevoort, and Y.~M. Asano, ``{VeRA}: Vector-based random
  matrix adaptation,'' in \emph{Proc. Int. Conf. Learn. Represent. (ICLR)},
  2024.

\bibitem{edalati2022krona}
A.~Edalati, M.~Tahaei, I.~Kobyzev, V.~P. Nia, J.~J. Clark, and
  M.~Rezagholizadeh, ``{KronA}: Parameter efficient tuning with {Kronecker}
  adapter,'' \emph{arXiv:2212.10650}, 2022.

\bibitem{wu2024mole}
X.~Wu, S.~Huang, and F.~Wei, ``Mixture of {LoRA} experts,'' in \emph{Proc. Int.
  Conf. Learn. Represent. (ICLR)}, 2024.

\bibitem{li2026flexilora}
Z.~Li \emph{et~al.}, ``{Flexi-LoRA} with input-adaptive ranks: Efficient
  finetuning for speech and reasoning tasks,'' \emph{arXiv:2605.01959}, 2026.

\bibitem{houlsby2019adapters}
N.~Houlsby \emph{et~al.}, ``Parameter-efficient transfer learning for {NLP},''
  in \emph{Proc. Int. Conf. Mach. Learn. (ICML)}, 2019.

\bibitem{li2018intrinsic}
C.~Li, H.~Farkhoor, R.~Liu, and J.~Yosinski, ``Measuring the intrinsic
  dimension of objective landscapes,'' in \emph{Proc. Int. Conf. Learn.
  Represent. (ICLR)}, 2018.

\bibitem{aghajanyan2021intrinsic}
A.~Aghajanyan, S.~Gupta, and L.~Zettlemoyer, ``Intrinsic dimensionality
  explains the effectiveness of language model fine-tuning,'' in \emph{Proc.
  59th Annu. Meeting Assoc. Comput. Linguistics (ACL-IJCNLP)}, 2021, pp.
  7319--7328.

\bibitem{valipour2023dylora}
M.~Valipour, M.~Rezagholizadeh, I.~Kobyzev, and A.~Ghodsi, ``{DyLoRA}:
  Parameter-efficient tuning of pre-trained models using dynamic search-free
  low-rank adaptation,'' in \emph{Proc. Conf. Eur. Chapter Assoc. Comput.
  Linguistics (EACL)}, 2023.

\bibitem{shinwari2025ardlora}
H.~U.~K. Shinwari and M.~Usama, ``{ARD-LoRA}: Dynamic rank allocation for
  parameter-efficient fine-tuning of foundation models with heterogeneous
  adaptation needs,'' \emph{IEEE Trans. Artif. Intell.}, 2025, early access.

\bibitem{shen2025kronlora}
Y.~Shen, ``{Kron-LoRA}: Hybrid {Kronecker}-{LoRA} adapters for scalable,
  sustainable fine-tuning,'' \emph{arXiv:2508.01961}, 2025.

\bibitem{bai2026cdka}
J.~Bai \emph{et~al.}, ``Diving into {Kronecker} adapters: Component design
  matters,'' in \emph{Proc. Int. Conf. Mach. Learn. (ICML)}, 2026.

\bibitem{sadeghi2025moka}
M.~Sadeghi \emph{et~al.}, ``{MoKA}: Mixture of {Kronecker} adapters,''
  \emph{arXiv:2508.03527}, 2025.

\bibitem{shazeer2017moe}
N.~Shazeer \emph{et~al.}, ``Outrageously large neural networks: The
  sparsely-gated mixture-of-experts layer,'' in \emph{Proc. Int. Conf. Learn.
  Represent. (ICLR)}, 2017.

\bibitem{fedus2022switch}
W.~Fedus, B.~Zoph, and N.~Shazeer, ``Switch transformers: Scaling to trillion
  parameter models with simple and efficient sparsity,'' \emph{J. Mach. Learn.
  Res.}, vol.~23, no. 120, pp. 1--39, 2022.

\bibitem{du2025moeosda}
Z.~Du, J.~An, Y.~Tu, J.~Hong, and D.~Wu, ``Mixture-of-experts for open set
  domain adaptation: A dual-space detection approach,'' \emph{IEEE Trans.
  Artif. Intell.}, vol.~6, no.~12, Dec. 2025.

\bibitem{jaradat2025hdp}
G.~A. Jaradat, M.~F. Tolba, G.~Alsuhli, H.~Saleh, M.~Al-Qutayri, and
  T.~Stouraitis, ``Efficient transformer inference through hybrid dynamic
  pruning,'' \emph{IEEE Trans. Artif. Intell.}, vol.~6, no.~12, Dec. 2025.

\bibitem{koohpayegani2024nola}
S.~A. Koohpayegani, K.~L. Navaneet, P.~Nooralinejad, S.~Kolouri, and
  H.~Pirsiavash, ``{NOLA}: Compressing {LoRA} using linear combination of
  random basis,'' in \emph{Proc. Int. Conf. Learn. Represent. (ICLR)}, 2024.

\bibitem{zhang2023lorafa}
L.~Zhang, L.~Zhang, S.~Shi, X.~Chu, and B.~Li, ``{LoRA-FA}: Memory-efficient
  low-rank adaptation for large language models fine-tuning,'' \emph{arXiv
  preprint arXiv:2308.03303}, 2023.

\bibitem{zhu2024asymmetry}
J.~Zhu \emph{et~al.}, ``Asymmetry in low-rank adapters of foundation models,''
  in \emph{Proc. Int. Conf. Mach. Learn. (ICML)}, ser. PMLR, vol. 235, 2024,
  pp. 62\,369--62\,385.

\bibitem{roy2007erank}
O.~Roy and M.~Vetterli, ``The effective rank: A measure of effective
  dimensionality,'' in \emph{Proc. Eur. Signal Process. Conf. (EUSIPCO)},
  Poznan, Poland, 2007, pp. 606--610.

\bibitem{rudelson2007sampling}
M.~Rudelson and R.~Vershynin, ``Sampling from large matrices: An approach
  through geometric functional analysis,'' \emph{J. ACM}, vol.~54, no.~4, p.
  Art. 21, 2007.

\bibitem{marchenko1967}
V.~A. Mar{\v{c}}enko and L.~A. Pastur, ``Distribution of eigenvalues for some
  sets of random matrices,'' \emph{Mathematics of the USSR-Sbornik}, vol.~1,
  no.~4, pp. 457--483, 1967.

\bibitem{yang2023medmnist}
J.~Yang \emph{et~al.}, ``{MedMNIST} v2: A large-scale lightweight benchmark for
  {2D} and {3D} biomedical image classification,'' \emph{Sci. Data}, vol.~10,
  no.~1, p.~41, 2023.

\bibitem{wang2019glue}
A.~Wang, A.~Singh, J.~Michael, F.~Hill, O.~Levy, and S.~R. Bowman, ``{GLUE}: A
  multi-task benchmark and analysis platform for natural language
  understanding,'' in \emph{Proc. Int. Conf. Learn. Represent. (ICLR)}, 2019.

\bibitem{anthropic2026claude}
{Anthropic}, ``{Claude} large language model,'' \url{https://claude.ai}, 2026.

\end{thebibliography}
\end{document}